\documentclass{article}
\usepackage[preprint]{neurips_2026}

\usepackage{amsmath,amssymb,amsfonts,amsthm,mathtools}

\usepackage[utf8]{inputenc}
\usepackage[T1]{fontenc}
\usepackage{hyperref}
\usepackage{url}
\usepackage{booktabs}
\usepackage{nicefrac}
\usepackage{microtype}
\usepackage{xcolor}
\usepackage{graphicx}
\usepackage{adjustbox}
\usepackage{subcaption}
\usepackage{multirow}
\usepackage{algorithm}
\usepackage{algorithmic}
\usepackage{enumitem}
\usepackage{wrapfig}
\usepackage{colortbl}

\usepackage[capitalize,noabbrev]{cleveref}

\setcitestyle{numbers,square,comma,sort&compress}

\theoremstyle{plain}

\theoremstyle{definition}

\theoremstyle{remark}

\crefname{assumption}{Assumption}{Assumptions}
\Crefname{assumption}{Assumptions}{Assumptions}

\definecolor{bestcolor}{HTML}{E8F5E9}
\definecolor{secondcolor}{HTML}{FFF3E0}
\definecolor{myblue}{HTML}{E3F2FD}

\makeatletter
\renewcommand\paragraph{\@startsection{paragraph}{4}{\z@}%
  {0.5ex \@plus 0.2ex \@minus 0.1ex}%
  {-0.5em}%
  {\normalfont\normalsize\bfseries}}
\renewcommand\subparagraph{\@startsection{subparagraph}{5}{\z@}%
  {0.5ex \@plus 0.2ex \@minus 0.1ex}%
  {-0.5em}%
  {\normalfont\normalsize\bfseries}}
\makeatother

\title{AnomalyClaw: A Universal Visual Anomaly Detection Agent via Tool-Grounded Refutation}

\author{%
  Xi Jiang$^{1}$\quad Yinjie Zhao$^{2,3}$\quad Zesheng Yang$^{1}$\quad Feng Zheng$^{1}$\thanks{Corresponding author.}\\[2pt]
  $^{1}$Department of Computer Science and Engineering, \\ Southern University of Science and Technology (SUSTech), Shenzhen, China\\
  $^{2}$School of EEE, Nanyang Technological University (NTU), Singapore \\[2pt]
  $^{3}$CFAR, Agency for Science, Technology and Research (A*STAR), Singapore\\[2pt]
  \texttt{jiangx2020@mail.sustech.edu.cn}\quad\texttt{f.zheng@ieee.org}
}

\begin{document}

\maketitle

\begin{abstract}

Visual anomaly detection (VAD) is crucial in many real-world fields, such as industrial inspection, medical imaging, infrastructure monitoring, and remote sensing. However, the specific anomaly definitions, data modalities, and annotation standards across different domains make it difficult to transfer single-domain trained VAD models. Vision-language models (VLMs), pre-trained on large-scale cross-domain data, can perform visual perception under task instructions, offering a promising solution for cross-domain VAD. However, single-inference VLM judgments are unreliable, since they rely more on prior knowledge than on normal-sample references or fine-grained feature evidence. We therefore present \textbf{AnomalyClaw}, a training-free VAD agent that turns anomaly judgment into a multi-round refutation process. In each round, the agent proposes candidate anomalies and refutes each against normal-sample references, drawing on a $13$-tool library for visual verification, reference parsing, and frozen expert probing. On the \textbf{CrossDomainVAD-12} benchmark ($12$ datasets), AnomalyClaw achieves consistent macro-AUROC improvements over single-step direct inference with $\mathbf{+6.23}$ pp on \textbf{GPT-5.5}, $\mathbf{+7.93}$ pp on \textbf{Seed2.0-lite}, and $\mathbf{+3.52}$ pp on \textbf{Qwen3.5-VL-27B}. We further introduce an optional \textbf{verbalized self-evolution} extension. It builds an online rulebook from internal-branch disagreement without oracle labels. On Qwen3.5-VL-27B, it delivers a $+2.09$ pp mean gain, comparable to a $K{=}10$ oracle-label supervised baseline ($+1.99$ pp). These results show that agentic refutation improve anomaly understanding and reasoning of VLMs, rather than merely aggregating tool outputs. We release the agent, the benchmark, and the analysis artifacts at \url{https://github.com/jam-cc/AnomalyClaw}.

\end{abstract}

\section{Introduction}
\label{sec:intro}

Visual anomaly detection (VAD) is widely used in real-world scenarios, including industrial defect detection~\citep{bergmann2019mvtec,zou2022visa}, medical lesion detection~\citep{baid2021brats,bao2024bmad}, road-scene safety identification~\citep{lis2019detecting}, infrastructure monitoring~\citep{dorafshan2018sdnet}, and remote-sensing change detection~\citep{chen2020levircd}. Each domain has its own anomaly definition and an independent distribution of normal images, and the labeled positive sample dataset is extremely small or even completely missing. Collecting large-scale, accurate anomaly annotation data remains the most costly aspect of deploying VAD. Furthermore, models trained in one type of scenario are difficult to transfer and adapt to other domains. Traditional anomaly detection algorithms cannot meet the demands of cross-domain applications. Therefore, a practical solution that requires no training, is ready to use out of the box, and can adapt to any new domain scenario is urgently needed.

\begin{figure}[t]
    \centering
    \includegraphics[width=1.0\linewidth]{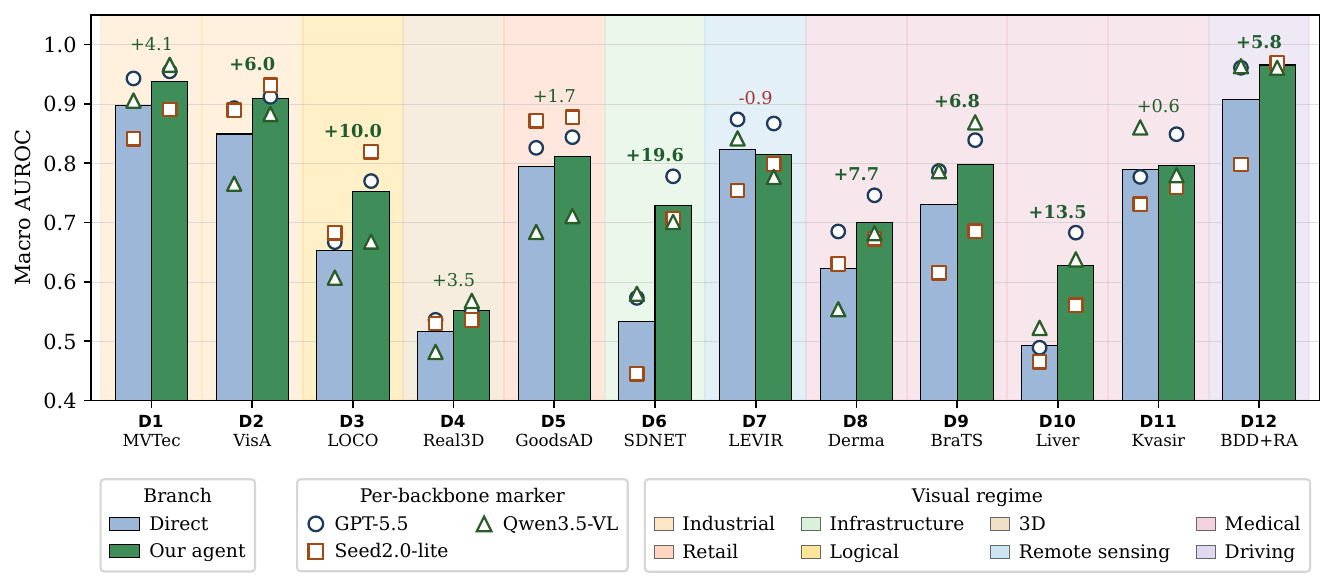}
    \vspace{-0.8em}
    \caption{\textbf{Cross-domain VAD is heterogeneous, and Direct prompting is uneven.} Per-domain macro AUROC on \textsc{CrossDomainVAD-12} test (D-codes in \S\ref{ssec:benchmark}). Bars are the mean over three VLM backbones; markers are per-backbone (GPT-5.5 $\circ$, Seed2.0-lite $\square$, Qwen3.5-VL-27B $\triangle$). Our agent (parallel Direct $+$ Refutation, $\alpha{=}0.5$) lifts the macro by $\Delta{=}+6.53$ pp.}
    \label{fig:motivation}
    \vspace{-1em}
\end{figure}

To this end, recent research has focused on exploring the generalization capabilities of multi-modal large models~\citep{jeong2023winclip,bao2024bmad,jiang2024mmad}, where a training-free vision-language model (VLM), inputting the query image and reference images, can directly output an anomaly score in a single call. However, general VLMs applied directly to anomaly detection still perform unsatisfactorily. Their built-in general prior knowledge conflicts with the anomaly definitions specific to anomaly detection tasks. For example, a fine crack is a serious defect on a road surface, but a natural, normal texture in a granite bedding surface; unfamiliar structures appearing between satellite image frames are precisely the terrain changes that need to be identified and labeled. Due to the lack of specific prior constraints on the morphology of \emph{normal samples} in various fields, VLMs often make judgments that seem to have high confidence but are actually wrong, and cannot be distinguished from correct judgments. Figure~\ref{fig:motivation} also shows that the detection performance of such models varies under different combinations of visual content and different backbone network selections.

We argue that VAD is essentially a combined perception task, requiring four sub-capabilities in sequence: locating candidate regions, comparing them with normal reference samples, applying domain-specific knowledge, and outputting a graded anomaly score, which beyond a single forward pass. The integrated approach is suitable for clean industrial benchmarks, but becomes insufficiently robust when the task requires explicit reference comparison or verification using domain-specific tools. Even if the model recognizes anomaly morphology in general, it still cannot go back to verify it. A reasonable improvement is to to decompose the forward pass into a multi-step tool-calling inference process~\citep{suris2023vipergpt,gupta2023visprog,yang2023mmreact,yao2023react,schick2023toolformer}, and the first batch of agent-based VAD systems~\citep{zhang2025agentiad,ji2025autoiad} has adopted this idea, building a loop that accumulates confirming evidence for the VLM's hypothesis. However, the core verification logic of VAD is inherently comparative. A candidate is anomalous only if it cannot be matched to any reference, yet accumulation-style loops tend to confirm rather than rule out. We therefore define each round of inference as an explicit \emph{refutation} check. The agent proposes candidate features, calls a few tools (side-by-side comparison, expert-score lookup, reference profiling, segmentation) to refute each against the normal references, and then scores.

We present \textbf{AnomalyClaw}, a training-free cross-domain VAD agent that operationalizes this decomposition. Each per-item invocation runs a $K{=}5$-turn refutation tool-using trajectory in which the agent proposes candidate features, invokes a \emph{specialty-aware} $13$-tool catalog, and disconfirms each candidate against the normal references. Per-tool applicability clauses keep $4$--$5$ tools in effective use per backbone, rather than collapsing onto a single primitive. The agent additionally runs a descriptor-free Direct VLM judgment on the \emph{same} backbone session and blends the two scores at a fixed $\alpha{=}0.5$ identical across backbones. On \textsc{CrossDomainVAD-12}, AnomalyClaw lifts macro AUROC over single-pass Direct by $+6.23$ pp on GPT-5.5, $+7.93$ pp on Seed2.0-lite, and $+3.52$ pp on Qwen3.5-VL-27B, all $95\%$-significant under stratified paired bootstrap.

Cross-domain VAD has no labeled-positive corpus by assumption, so any improvement loop must be self-supervised by construction. The dominant LLM-agent self-improvement recipes~\citep{shinn2023reflexion,zhao2024expel,wang2023voyager} all depend on an external success signal such as a reward, an executable oracle, or a ground-truth label, none of which we can construct here. We instead turn the agent's own internal branch disagreement into that supervisory signal. Whenever the Direct and refutation branches disagree by more than $0.30$ on an item, the item enters a per-domain queue. Once the queue fills, a reflector VLM reads the accumulated batch and writes corrective textual rules tagged by domain and category, and subsequent items retrieve those rules via metadata RAG before running. The signal exists \emph{only} because the system has two branches to disagree with itself in, so disagreement-based feedback applies only to an agent like ours rather than to any VLM call, making the upgrade an agent-native capability rather than a generic prompt trick. The full design is verbalized throughout, with no parameter updates and no oracle labels at any point. This optional \textbf{verbalized self-evolution} extension delivers a $+2.09$ pp mean improvement on the weakest backbone Qwen3.5-VL-27B at $P(\Delta_{\text{boot}}{>}0){=}0.87$, where the bootstrap CI crosses zero. The mean gain matches a $K{=}10$-oracle-label rule-learning baseline at $+1.99$ pp and exceeds a same-budget $\alpha$-tuning baseline at $-0.21$ pp.

\paragraph{Contributions.}
\textbf{(1) The first universal VAD agent.} To our knowledge, AnomalyClaw is the first training-free, backbone-agnostic VAD agent. It frames the task as compositional perception and casts each turn as an explicit \emph{refutation} of candidate features against the normal references, using a specialty-aware tool catalog, with the same recipe running unchanged across VLM backbones.
\textbf{(2) Verbalized self-evolution from branch disagreement.} The agent grows its own rulebook online during a single test pass with zero oracle labels, an agent-native capability that label-based loops cannot replicate.
\textbf{(3) Empirical and behavioural analysis.} AnomalyClaw lifts macro AUROC over single-pass Direct on three diverse backbones with $95\%$-significant gains everywhere, and the accompanying per-backbone diagnostics on reasoning depth, tool usage, and refutation-verdict distribution are runtime-observable and inform deployment triage.
\textbf{(4) \textsc{CrossDomainVAD-12}}, a $12$-domain reference-based benchmark spanning industrial, retail, logical, 3D, infrastructure, remote-sensing, medical, and road-scene sources.

\section{Related Work}
\label{sec:related_work}

\paragraph{Frozen-backbone single-pass VAD.}
A long line of VAD work scores query patches against frozen visual representations~\citep{jiang2022vadsurvey,roth2022patchcore,jiang2022softpatch,wang2025softpatchplus,jeong2023winclip,zhou2024anomalyclip,damm2025anomalydino,subspacead2026,jiang2024multiclass,anomalyvfm2026,hou2026visualad}. Some are training-free at deployment (memory banks: PatchCore~\citep{roth2022patchcore}, AnomalyDINO~\citep{damm2025anomalydino}; PCA subspaces: SubspaceAD~\citep{subspacead2026}); others learn lightweight auxiliary modules off-target while keeping the target domain access zero-shot (object-agnostic prompts: AnomalyCLIP~\citep{zhou2024anomalyclip}; LoRA on a vision foundation model: AnomalyVFM~\citep{anomalyvfm2026}; learnable language-free token heads: VisualAD~\citep{hou2026visualad}). All of them assume the anomaly signal is visible in a generic embedding without any reasoning step. A parallel thread fine-tunes or evaluates large VLMs directly on the task~\citep{gu2024anomalygpt,jiang2026adcopilot,jiang2024mmad,bao2024bmad}. The dominant deployment pattern across both threads is a \emph{single forward pass} with one or two references---no per-image evidence gathering, and no chance for the VLM to challenge its own first impression.

\paragraph{Agentic anomaly detection.}
A recent thread of agentic VAD systems extends the single-pass VLM call with structured multi-step iteration, but the iteration is \emph{baked into model weights through training}, not realized at prompt time. AgentIAD~\citep{zhang2025agentiad} couples a VLM with a perceptive zoomer and a comparative retriever, trained via supervised fine-tuning plus agentic reinforcement learning, so the tool-use schedule and multi-turn loop live inside the fine-tuned weights, meaning a different backbone or a new tool requires re-training. AutoIAD~\citep{ji2025autoiad} orchestrates specialist sub-agents through a manager that is itself fine-tuned on industrial data. EAGLE~\citep{peng2026eagle} steers the VLM via threshold-guided prompt selection and confidence-aware attention sharpening over an expert-detector signal, an attention-guidance recipe rather than a tool-using loop. All three evaluate only against industrial benchmarks and adopt an \emph{accumulation} stance, where each turn or guidance signal adds more evidence to the VLM's working hypothesis. By contrast, we run the multi-turn tool loop entirely at prompt time and cast each turn as an explicit \emph{refutation} of the working hypothesis, putting AnomalyClaw conceptually closer to general perception agents like ViperGPT~\citep{suris2023vipergpt}, VisProg~\citep{gupta2023visprog}, and MM-ReAct~\citep{yang2023mmreact} and the ReAct/Toolformer line~\citep{yao2023react,schick2023toolformer} than to any of the systems above.

\paragraph{Verbalized self-improvement in LLM agents.}
A separate line of agent research shows that an LLM can improve by reading and writing natural-language lessons rather than running parameter updates. Reflexion~\citep{shinn2023reflexion} introduces ``verbal reinforcement learning'' from failed trajectories, ExpeL~\citep{zhao2024expel} extracts reusable rules across trajectories by contrasting successes and failures, and Voyager~\citep{wang2023voyager} stores analogous lessons as skill code. All three rely on an \emph{external} supervisory signal---reward, unit test, or oracle label---which cross-domain VAD does not provide. AnomalyClaw instead drives the same recipe from the agent's own \emph{internal} branch disagreement. Self-refine~\citep{madaan2023selfrefine} and multi-agent debate~\citep{du2023debate,irving2018debate} refine \emph{within} a single instance instead, and layered unconditionally on a strong descriptor-enhanced baseline, they regress it by $-2.4$ to $-10.2$ pp in our setting, which motivated the cross-instance verbalized-learning route. Appendix~\ref{app:failed_variants} reports the full set of failed variants.

\paragraph{Cross-domain visual anomaly detection.}
Most VAD literature is anchored to a single application domain. Industrial inspection is by far the most developed thread~\citep{bergmann2019mvtec,zou2022visa,bergmann2022mvtecloco}; retail-product anomaly detection~\citep{zhang2024goodsad}, civil-infrastructure crack inspection~\citep{dorafshan2018sdnet}, medical anomaly detection~\citep{baid2021brats,borgli2020hyperkvasir,yang2023medmnist,bao2024bmad}, remote-sensing change detection~\citep{chen2020levircd}, and road-anomaly detection~\citep{lis2019detecting} (where broader driving datasets such as BDD100K~\citep{yu2020bdd100k} are repurposed for the same task) have each developed in isolation, with detectors and protocols tuned per task. MMAD~\citep{jiang2024mmad} broadens evaluation across industrial sub-domains but still restricts itself to multiple-choice QA. CrossDomainVAD-12 unifies $12$ distinct domains into a single image-level reference-based AUROC protocol with frozen splits, exposing backbone- and domain-dependent failure modes that single-domain evaluation masks.

\section{Method}
\label{sec:method}

\subsection{Overview}
\label{ssec:overview}

\begin{figure}[t]
    \centering
    \adjustbox{center}{\includegraphics[width=1.05\linewidth]{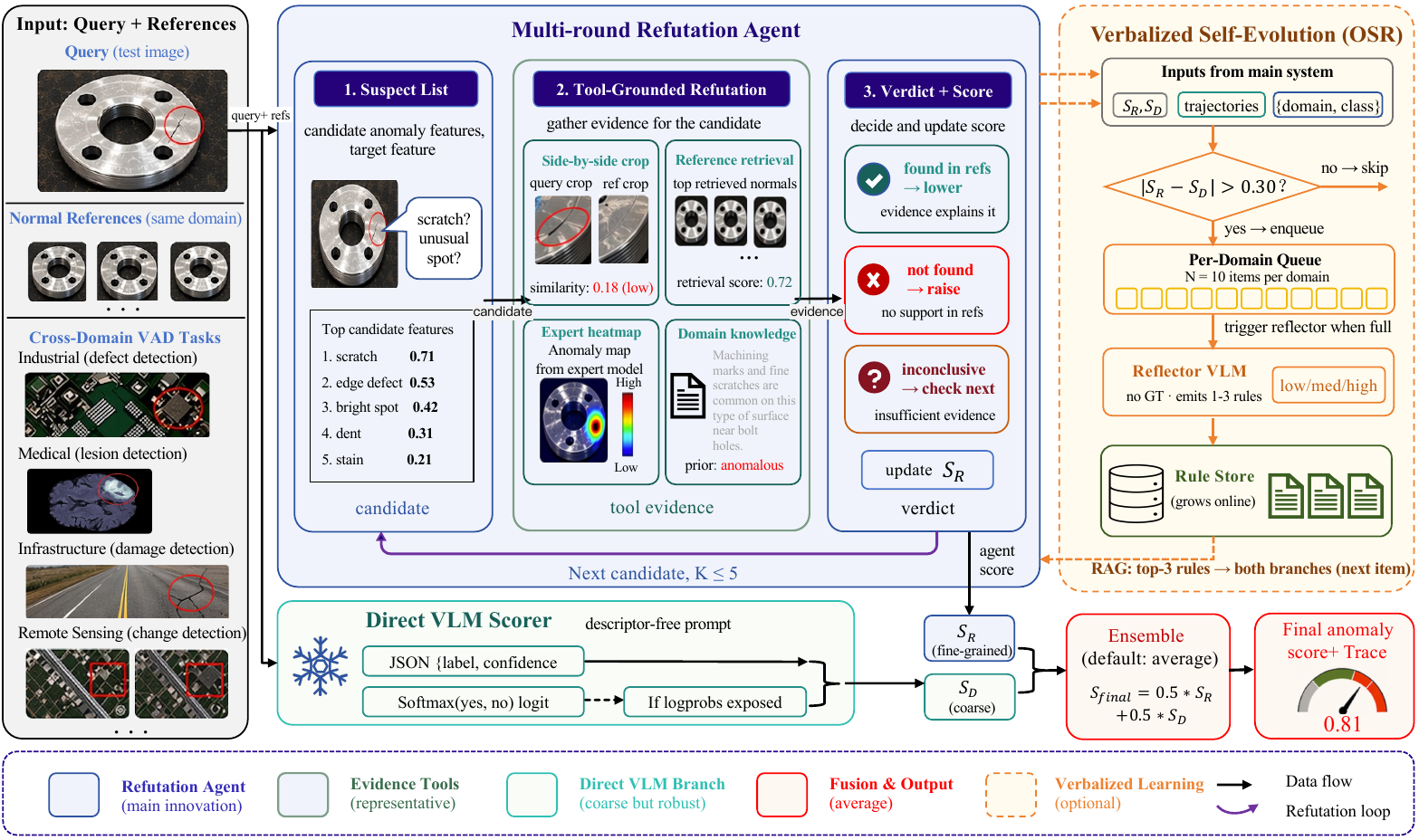}}
    \caption{\textbf{AnomalyClaw architecture.} The system is centred on a multi-turn \textbf{refutation agent} (main innovation): each turn cycles through a \emph{suspect list} of candidate features, \emph{tool-grounded refutation evidence} from a $13$-entry specialty-aware tool catalog, and a \emph{verdict and score update}. A descriptor-free \textbf{Direct VLM} call on the same backbone provides a complementary coarse score; the two are averaged at a fixed $\alpha{=}0.5$ (no dev tuning, identical across all three backbones). The orange panel on the right is an \emph{optional} verbalized-self-evolution loop (\S\ref{ssec:osr}) that grows a domain rulebook online from internal branch disagreement with zero oracle labels.}
    \label{fig:architecture}
    \vspace{-1em}
\end{figure}

For each test item, AnomalyClaw runs three components on the same backbone API session (Figure~\ref{fig:architecture}): a multi-turn refutation agent, a single-call Direct VLM scorer, and an optional verbalized self-evolution loop. The refutation agent is the system's main innovation and accounts for most of its behavioral signal; the Direct scorer is a coarse but robust co-signal on the same backbone; verbalized self-evolution is detailed separately in \S\ref{ssec:osr}.

\paragraph{Refutation agent (main).}
A structured multi-turn trajectory (\S\ref{ssec:refutation}) with a $K{=}5$ turn budget. Each turn cycles through three stages: maintain a \emph{suspect list} of candidate anomalous features, gather \emph{tool-grounded refutation evidence} from a $13$-entry specialty-aware tool catalog (visual inspection, reference retrieval, expert heatmap, domain-knowledge lookup, etc.), and emit a \emph{verdict and updated score}. Output: refuted anomaly score $s_R \in [0,1]$.

\paragraph{Direct VLM scorer (co-signal).}
A single descriptor-free VLM call on the same backbone, in either of two interchangeable forms: a JSON \verb|{label, confidence}| readout, or a softmax over the first generated token's $\log p(\text{yes})$ and $\log p(\text{no})$ when the inference engine exposes logprobs (\S\ref{ssec:logit_direct}). Both forms are coarse compared to the agent and serve as an independent reading from the same backbone. Output: scalar score $s_D \in [0,1]$.

\paragraph{Fusion.}
$s_{\mathrm{final}} = 0.5\,s_D + 0.5\,s_R$, fixed a priori with no dev tuning and identical across backbones. The two components issue concurrently on the same session: wall-time is $\max(\text{Direct},\text{agent})$ rather than the sum, but token-billed cost is the sum --- AnomalyClaw averages $3.07$--$3.42$ VLM calls per item across backbones against $1$ for single-pass Direct, reaching $5$--$8$ in the long tail on the weakest backbone. Per-backbone, per-domain, and per-tool budget breakdowns are in Appendix~\ref{app:compute_budget}.

\subsection{Refutation protocol}
\label{ssec:refutation}

The refutation agent targets two failure modes of unstructured ReAct-style loops. The first is a \emph{coarse rank grid} that caps ranking resolution. The second is \emph{confirmation bias}, where tool calls support the VLM's working hypothesis rather than test it. Each turn cycles through the three stages below.

\textit{(1) Suspect list.} Maintain up to three \texttt{candidate\_features} with per-feature suspicion scores, and pick a \texttt{refutation\_target} for the current turn. Turn $1$ is tool-free and seeds this list together with an \texttt{initial\_score}; if the list is empty and the initial score is $<0.3$, the item finalizes early at $0.05$.

\textit{(2) Tool-grounded refutation.} The agent picks one tool per turn from a specialty-aware \textbf{L1--L7 ladder} ordered by cost rather than capability.
\begin{itemize}[leftmargin=*,topsep=2pt,itemsep=0pt]
\item L1 \texttt{reference\_profiler} --- text-level ``what is normal here?'' summary;
\item L2 \texttt{side\_by\_side(bbox)} --- visual query-vs-refs at a region;
\item L3 \texttt{expert\_score} --- cached pretrained AD model + heatmap + suggested\_bbox;
\item L4 \texttt{reference\_retriever} --- pull additional normal images via DINOv2~\citep{oquab2024dinov2} similarity;
\item L5 \texttt{zoom\_bbox} --- single-image high-res crop of the query;
\item L6 structural analytics --- \texttt{segment\_and\_count}, \texttt{component\_counter}, \texttt{texture\_fft};
\item L7 \texttt{image\_diff} / \texttt{rotate\_align} --- pixel-level diff for aligned industrial domains only.
\end{itemize}
The system prompt instructs the agent to pick the lowest rung that resolves the current uncertainty. Tool applicability is hard-gated by domain. \texttt{image\_diff} and \texttt{rotate\_align} refuse on non-aligned domains such as medical, aerial, and road, and \texttt{expert\_score} returns \texttt{available=False} on D2, D4, D8, D11, and D12, where the cached expert AUROC is unreliable. Each item also receives a \emph{per-domain hint} as part of its turn-$1$ user message, naming the domain's expert reliability and the recommended tool pipeline; empirically, this hint drives $4$--$5$ tools into effective use per backbone, rather than collapsing onto a single primitive. Tool observations are paired with an explicit disconfirmatory clause that biases use toward \emph{ruling out anomalies}.

\textit{(3) Verdict and score update.} The VLM emits a \texttt{refutation\_verdict} from \texttt{found\_in\_ref}, \texttt{not\_found}, or \texttt{inconclusive}, removes refuted features, and writes an \texttt{updated\_score} clamped to $[0.05,0.20]$ when all features are refuted or $[0.40,0.95]$ when any survive. If the trajectory has not been finalized by $K{=}5$, a \emph{forced-final} message requires the agent to commit on its current evidence. Appendices~\ref{app:specialty} and~\ref{app:refutation} contain the JSON schema, per-tool contracts, and task-preamble builder.

\subsection{Direct scorer}
\label{ssec:logit_direct}

The default Direct scorer asks the VLM for a JSON object \verb|{image_label, confidence}| and reads $s_D{=}$\verb|confidence| if \verb|image_label==anomalous| else $1{-}$\verb|confidence|. When the inference engine exposes first-token logprobs (e.g.\ vLLM), we replace the prompt with a yes/no question and read $s_D^{\mathrm{logit}} = e^{\ell_y}/(e^{\ell_y}{+}e^{\ell_n})$, a logit-derived proxy probability for $P(\text{anomalous})$ before tokenisation rounds it to a hard label; this gives a finer rank grid from the same single call. Of our three backbones, only Qwen3.5-VL-27B exposes logprobs and uses the logit form; GPT-5.5 and Seed2.0-lite use the JSON form. Both forms are coarse compared to the refutation agent's score, with ablations and unique-value statistics in Appendix~\ref{app:mechanism}.

\subsection{Verbalized self-evolution (optional)}
\label{ssec:osr}

The main agent (\S\ref{ssec:overview}) re-reads the references on every query but never \emph{accumulates} anything across them, so identical refs give identical traces. As an optional add-on, we couple the same agent with \textbf{Online Self-Refinement (OSR)}, a zero-oracle test-time loop that lets the agent grow its own rulebook from \emph{internal branch disagreement}. OSR is orthogonal to the parallel-Direct $+$ refutation core (\S\ref{ssec:overview}--\S\ref{ssec:logit_direct}); when disabled, the system reduces exactly to the always-on agent.

\paragraph{Algorithm.} The rulebook starts empty. For each test item:
\begin{enumerate}[leftmargin=*,topsep=2pt,itemsep=0pt]
\item \textbf{Retrieve.} Top-$3$ rules filtered by the query's $(\text{domain}, \text{category})$ metadata are concatenated and prepended as user-message context to \emph{both} the Direct prompt and the refutation turn-$1$ message.
\item \textbf{Run.} The main agent (\S\ref{ssec:overview}) emits $s_{D}$, $s_{R}$, and $s_{\text{final}}{=}\alpha s_D + (1{-}\alpha) s_R$.
\item \textbf{Detect disagreement.} If $|s_D - s_R| > \tau_d{=}0.30$, push the item (with both branches' trajectories) into a per-domain queue.
\item \textbf{Reflect.} When the per-domain queue reaches $N_{\text{batch}}{=}10$, a reflector VLM reads the batch (\emph{no GT labels}), proposes $1$--$3$ corrective rules with a self-assigned confidence, and rules at $\ge$ medium confidence enter the store, tagged by source category.
\end{enumerate}

\paragraph{Hyperparameters.} $\tau_d{=}0.30$ is calibrated against the \S\ref{ssec:overview} branch-difference distribution (Direct's hard-binary output and the refutation's $[0.05,0.20]\cup[0.40,0.95]$ clip jointly produce active mass around $0.10$--$0.30$); $N_{\text{batch}}{=}10$ is fixed a priori. The blend weight $\alpha$ is the same $0.5$ as the \S\ref{ssec:overview} main agent --- OSR introduces no tunable knob beyond the rule store.

\paragraph{What is not learned.} OSR's reflector sees query images and references, but is never told the per-query anomaly label and never receives a gradient. The mechanism is a textual rule-store update with no parameter changes anywhere.

\section{Experiments}
\label{sec:main}

\subsection{Benchmark: CrossDomainVAD-12}
\label{ssec:benchmark}

We evaluate on \textsc{CrossDomainVAD-12}, a twelve-domain reference-based anomaly-detection benchmark under a single per-image AUROC protocol: D1 MVTec-AD industrial~\citep{bergmann2019mvtec}, D2 VisA complex industrial~\citep{zou2022visa}, D3 MVTec-LOCO logical~\citep{bergmann2022mvtecloco}, D4 Real3D-AD 3D product~\citep{liu2023real3dad} (rendered point-cloud views), D5 GoodsAD retail~\citep{zhang2024goodsad}, D6 SDNET2018 concrete infrastructure~\citep{dorafshan2018sdnet}, D7 LEVIR-CD+ bi-temporal remote-sensing change~\citep{chen2020levircd}, D8 DermaMNIST dermoscopy~\citep{yang2023medmnist}, D9 BraTS2021 brain MRI~\citep{baid2021brats}, D10 BMAD-Liver CT~\citep{bao2024bmad}, D11 HyperKvasir GI endoscopy~\citep{borgli2020hyperkvasir}, D12 BDD100K+RoadAnomaly21 road safety~\citep{yu2020bdd100k,chan2021segmentmeifyoucan,lis2019detecting}. Each domain contributes $20$ calibration / $40$ dev / $120$ test items (D7: $98$ test; LEVIR-CD+ contributes $158$ bi-temporal pairs total). All items are reference-based and paired with $1$--$10$ normal references. Total test items $n{=}1{,}418$. Per-domain specifications, sampling, descriptor sentences, and licenses are in Appendix~\ref{app:benchmark}.

\subsection{Setup}
\label{ssec:setup}

Three VLMs spanning proprietary--open: \textbf{GPT-5.5} (proprietary frontier), \textbf{Seed2.0-lite} (\texttt{doubao-seed-2-0-lite}, proprietary non-frontier), and \textbf{Qwen3.5-VL-27B} (open-weight, served locally via vLLM~\citep{kwon2023vllm} with thinking disabled, exposing first-token logprobs that enable our logit-as-soft-label Direct branch; the GPT-5.5 and Seed2.0-lite chat-completions endpoints we use do not expose logprobs and so fall back to JSON-confidence Direct). Temperature $0$, identical prompts across backbones. We report per-domain AUROC and macro-averaged across the 12 test domain codes; all $95\%$ confidence intervals are stratified paired bootstraps with per-domain stratification and $1{,}000$ resamples (Appendix~\ref{app:bootstrap}).

\subsection{Per-domain results}
\label{ssec:three_regimes}

\begin{table}[t]
\centering
\caption{\textbf{Per-domain AUROC on CrossDomainVAD-12 test}. The default protocol is a $4$-shot reference, and $0$-shot variants are marked. Macro $\Delta$ is paired bootstrap with $\star$ marking $95\%$ significance. \textbf{Bold} marks the larger of Direct vs Ours within each backbone pair, per column.}
\label{tab:main_results}
\scriptsize
\setlength{\tabcolsep}{2.8pt}
\resizebox{\linewidth}{!}{%
\begin{tabular}{l|cccccccccccc|c}
\toprule
Method & D1 & D2 & D3 & D4 & D5 & D6 & D7 & D8 & D9 & D10 & D11 & D12 & \textbf{Macro}\;($\Delta$) \\
\midrule
\multicolumn{14}{l}{\emph{General VLM}} \\
\hspace{1mm}GPT-5.5         & 0.943 & 0.893 & 0.667 & 0.536 & 0.826 & 0.573 & \textbf{0.874} & 0.685 & 0.787 & 0.489 & 0.777 & 0.961 & 0.751 \\
\hspace{1mm}Seed2.0-lite          & 0.841 & 0.889 & 0.683 & 0.530 & 0.872 & 0.445 & 0.754 & 0.630 & 0.616 & 0.465 & 0.731 & 0.798 & 0.688 \\
\hspace{1mm}Qwen3.5-VL-27B (JSON) & 0.906 & 0.766 & 0.607 & 0.482 & \textbf{0.684} & 0.580 & 0.842 & 0.554 & 0.787 & 0.522 & \textbf{0.860} & 0.964 & 0.713 \\
\midrule
\multicolumn{14}{l}{\emph{Our agent}} \\
\hspace{1mm}GPT-5.5         & \textbf{0.955} & \textbf{0.912} & \textbf{0.770} & \textbf{0.550} & \textbf{0.844} & \textbf{0.778} & 0.867 & \textbf{0.746} & \textbf{0.839} & \textbf{0.683} & \textbf{0.849} & \textbf{0.967} & \textbf{0.814}\;($+6.23^{\star}$) \\
\hspace{1mm}Seed2.0-lite          & \textbf{0.891} & \textbf{0.932} & \textbf{0.819} & \textbf{0.536} & \textbf{0.878} & \textbf{0.706} & \textbf{0.799} & \textbf{0.673} & \textbf{0.685} & \textbf{0.560} & \textbf{0.759} & \textbf{0.969} & \textbf{0.767}\;($+7.93^{\star}$) \\
\hspace{1mm}Qwen3.5-VL-27B  & \textbf{0.956} & \textbf{0.773} & \textbf{0.673} & \textbf{0.517} & 0.649 & \textbf{0.672} & 0.842 & \textbf{0.630} & \textbf{0.841} & \textbf{0.616} & 0.817 & \textbf{0.990} & \textbf{0.748}\;($+3.52^{\star}$) \\
\midrule
\multicolumn{14}{l}{\emph{Specialist non-VLM AD methods}} \\
\hspace{1mm}SubspaceAD~\citep{subspacead2026}        & 0.966 & 0.919 & 0.701 & 0.565 & 0.841 & 0.757 & 0.566 & 0.688 & 0.816 & 0.710 & 0.608 & 0.989 & 0.760 \\
\hspace{1mm}AnomalyDINO~\citep{damm2025anomalydino}  & 0.975 & 0.889 & 0.642 & 0.530 & 0.842 & 0.727 & 0.687 & 0.695 & 0.739 & 0.566 & 0.635 & 0.982 & 0.742 \\
\hspace{1mm}VisualAD~\citep{hou2026visualad}(0-shot)         & 0.882 & 0.936 & 0.594 & 0.497 & 0.704 & 0.603 & 0.345 & 0.730 & 0.689 & 0.675 & 0.547 & 0.725 & 0.661 \\
\hspace{1mm}AnomalyVFM~\citep{anomalyvfm2026}(0-shot)        & 0.890 & 0.736 & 0.572 & 0.478 & 0.566 & 0.798 & 0.306 & 0.572 & 0.791 & 0.577 & 0.540 & 0.443 & 0.606 \\
\midrule
\multicolumn{14}{l}{\emph{Specialist VLM}} \\
\hspace{1mm}AD-Copilot~\citep{jiang2026adcopilot}   & 0.961 & 0.873 & 0.574 & 0.539 & 0.692 & 0.688 & 0.589 & 0.657 & 0.712 & 0.573 & 0.748 & 0.793 & 0.700 \\
\hspace{1mm}IAD-R1~\citep{li2025iadr1} (0-shot)             & 0.811 & 0.782 & 0.477 & 0.496 & 0.587 & 0.650 & 0.230 & 0.505 & 0.470 & 0.752 & 0.449 & 0.214 & 0.535 \\
\bottomrule
\end{tabular}%
}
    \vspace{-0.8em}
\end{table}

Table~\ref{tab:main_results} and Figure~\ref{fig:hero_appendix} report AnomalyClaw's per-domain AUROC on all three backbones, fusing the refutation branch with the \emph{same} JSON-confidence Direct form as the baseline. The macro gain over JSON Direct is significant at $95\%$ on every backbone: $+6.23$ pp on GPT-5.5 ($95\%$ CI $[+4.84, +7.63]$), $+7.93$ pp on Seed2.0-lite ($[+6.20, +9.53]$), and $+3.52$ pp on Qwen3.5-VL-27B ($[+1.93, +5.11]$); all $P(\Delta{>}0){>}0.999$ ($0/1{,}000$ resamples $\le 0$). Since Direct and Ours use the same JSON-confidence form on Qwen, the $+3.52$ pp gain isolates the refutation agent from any score-extraction effect. An orthogonal optimisation reading Qwen's vLLM logprobs (\S\ref{ssec:logit_direct}) lifts Direct alone from $0.713$ to $0.764$ (pure score-extraction, no agent); fusing the agent with this logit-Direct reaches $0.767$ in deployment. Appendix~\ref{app:logit_direct_ablation} attributes each contribution independently. Removing Direct and keeping only refutation (Appendix~\ref{app:branch_ablation}), the agent still improves over JSON Direct on $32/36$ domain~$\times$~backbone cells, with negative cells confined to four $(D, B)$ pairs ($\Delta\in[-4.3, -0.7]$ pp) on Qwen3.5-VL-27B or GPT-5.5 D7.

\paragraph{Comparison to non-VLM and specialist VLM baselines.}
SubspaceAD~\citep{subspacead2026} is the strongest non-VLM training-free baseline at macro $0.760$; our agent (JSON-fusion as in Table~\ref{tab:main_results}) matches or exceeds it on GPT-5.5 ($0.814$) and Seed2.0-lite ($0.767$) and is within $1.2$ pp on Qwen3.5-VL-27B ($0.748$). Specialist VLMs underperform the base they were fine-tuned from on cross-domain evaluation. The base Qwen2.5-VL-7B-Instruct~\citep{bai2025qwen25vl}, scored with the same $4$-shot Yes/No-logit protocol, reaches $0.738$ macro AUROC; AD-Copilot~\citep{jiang2026adcopilot} (same base $+$ SFT on Chat-AD $+$ comparison encoder) drops to $0.700$ ($-3.8$ pp), and IAD-R1~\citep{li2025iadr1} (same base $+$ GRPO on Expert-AD CoT) drops to $0.535$ in its official $0$-shot mode ($-20.3$ pp), collapsing below random on D7 LEVIR ($0.230$) and D12 BDD-RA ($0.214$). The gap concentrates outside MVTec-AD: AD-Copilot loses $-18.5$ pp on D3 logical, $-21.7$ pp on D7 LEVIR, and $-9.9$ pp on D8 dermoscopy versus the base, and IAD-R1 loses on \emph{every} non-MVTec domain. Industrial-domain fine-tuning erodes the base VLM's cross-domain generalization, while our agent on Qwen3.5-VL-27B reaches $0.748$ (JSON-fusion) without task-specific training. The biggest agent-vs-baseline gaps are on \textbf{D7} LEVIR (non-VLM $\le 0.69$ vs ours $0.80$--$0.84$), \textbf{D11} Kvasir endoscopy (AD-Copilot $0.748$ vs ours $0.76$--$0.82$), and \textbf{D3} MVTec-LOCO logical (all baselines $\le 0.76$ vs ours up to $0.82$).

\paragraph{Where the gain concentrates.}
The lift over Direct is not uniform across domains (Figure~\ref{fig:per_domain_appendix}). The largest backbone-mean gains concentrate where Direct is weakest: D6 SDNET cracks ($+19.6$ pp), D10 BMAD-Liver CT ($+13.5$ pp), D3 MVTec-LOCO logical ($+10.0$ pp), D8 DermaMNIST ($+7.7$ pp); on easier domains where Direct already exceeds $0.85$ on every backbone (D1, D7, D12) the gain shrinks to $+1$--$6$ pp. Two domains regress on the per-backbone mean (D5 GoodsAD $-0.5$, D7 LEVIR $-0.9$), each driven by one or two negative-cell backbones balanced against positive ones; these residual failures motivate the verbalized self-evolution extension in \S\ref{ssec:osr_results}.

\subsection{Cross-task generalization: anomaly-related QA (MMAD)}
\label{ssec:mmad}

We further evaluate AnomalyClaw on MMAD~\citep{jiang2024mmad}, the
established benchmark for VLM-based industrial anomaly reasoning,
which contains $39{,}672$ multiple-choice questions across seven
subtasks (Anomaly Discrimination, Defect
Classification/Localization/Description/Analysis, Object
Classification/Analysis) and four
source datasets (DS-MVTec, MVTec-LOCO, VisA, GoodsAD). We follow the
standard $1$-shot setting and report letter accuracy on a stratified
$n{=}500$-image dev sample ($n{=}2{,}302$ questions) with seed $=42$,
identical to MMAD's release protocol. AnomalyClaw deploys two
backbones --- the open-source Qwen3.5-VL-27B and the proprietary
GPT-5.5 --- and applies a unified task-aware agent that auto-detects
The question type retrieves four DINOv2-similar normal references
from the per-domain bank in turn~$1$, then issues an MCQ-style letter
answer (option-score argmax) for AD or runs the refutation pipeline
elsewhere. Crucially, AnomalyClaw is \emph{training-free}: no MMAD
data is used to fit any parameter.

\begin{table*}[t]
\centering
\caption{\textbf{Performance comparison on MMAD with the standard
$1$-shot setting} (letter accuracy \%, averaged over the four source
datasets). \textbf{Best} and \underline{second-best} per column are marked among general MLLMs only; AD-specialist methods (gray) are shown for reference.}
\label{tab:mmad}
\scriptsize
\setlength{\tabcolsep}{3pt}
\resizebox{\linewidth}{!}{%
\begin{tabular}{c|c|c|cccc|cc|c}
\toprule
                         &                         & Anomaly        & \multicolumn{4}{c|}{Defect}                             & \multicolumn{2}{c|}{Object} &                           \\\cmidrule(r){3-9}
\multirow{-2}{*}{Model} & \multirow{-2}{*}{Scale} & Discrimination & Classification & Localization & Description & Analysis & Classification  & Analysis & \multirow{-2}{*}{Average} \\\midrule
\rowcolor{myblue}
Human (expert)          & -                       & 95.24          & 75.00          & 92.31        & 83.33       & 94.20    & 86.11           & 80.37    & 86.65 \\
\rowcolor{myblue}
Human (ordinary)        & -                       & 86.90          & 66.25          & 85.58        & 71.25       & 81.52    & 89.58           & 69.72    & 78.69 \\
\midrule
InternVL2~\citep{chen2024internvl2}               & 76B  & 68.25          & 54.22          & 56.66        & 66.30       & 80.47    & 86.40           & 82.92    & 70.75 \\
Qwen2.5-VL~\citep{bai2025qwen25vl}              & 72B  & \textbf{72.66}          & 62.31          & \underline{67.16}        & 73.56       & 81.95    & 94.30           & \underline{86.78}    & \underline{76.96} \\
Qwen3-VL                & 8B   & 68.71          & 63.27          & 61.01        & 68.18       & 79.99    & 92.27           & 83.96    & 73.91 \\
\midrule
Gemini-1.5-pro~\citep{team2024gemini15}          & -    & 68.63          & 60.12          & 58.56        & 70.38       & 82.46    & 89.20           & 82.25    & 73.09 \\
Seed1.5-VL~\citep{team2025seed15vl}              & -    & 65.30          & \underline{64.32}          & 63.35        & 73.80       & \underline{84.38}    & 91.65           & 83.67    & 75.21 \\
GPT-5.5       & -    & 69.95          & \textbf{64.91}          & 62.34        & \underline{75.95}       & 80.10    & \underline{95.23}           & 82.11    & 76.20 \\
\textbf{AnomalyClaw (GPT-5.5)} & -    & \underline{72.51}          & 63.64          & \textbf{67.73}        & \textbf{79.65}       & \textbf{87.99}    & \textbf{95.38}           & \textbf{87.13}    & \textbf{79.15} \\
\midrule
\textcolor{gray}{AnomalyGPT}        & \textcolor{gray}{7B}   & \textcolor{gray}{65.57}          & \textcolor{gray}{27.49}          & \textcolor{gray}{27.97}        & \textcolor{gray}{36.86}       & \textcolor{gray}{32.11}    & \textcolor{gray}{29.84}           & \textcolor{gray}{35.82}    & \textcolor{gray}{36.52} \\
\textcolor{gray}{EIAD~\citep{zhang2025eiad}}             & \textcolor{gray}{8B}   & \textcolor{gray}{60.50}          & \textcolor{gray}{50.70}          & \textcolor{gray}{55.60}        & \textcolor{gray}{67.80}       & \textcolor{gray}{76.90}    & \textcolor{gray}{91.80}           & \textcolor{gray}{82.70}    & \textcolor{gray}{69.40} \\
\textcolor{gray}{AgentIAD}              & \textcolor{gray}{3B}   & \textcolor{gray}{69.49}  & \textcolor{gray}{72.73}  & \textcolor{gray}{80.94}  & \textcolor{gray}{85.27}  & \textcolor{gray}{87.84}  & \textcolor{gray}{93.27}  & \textcolor{gray}{90.59}  & \textcolor{gray}{82.88} \\
\textcolor{gray}{AD-Copilot-thinking}  & \textcolor{gray}{7B}   & \textcolor{gray}{73.95}  & \textcolor{gray}{74.29}  & \textcolor{gray}{76.40}  & \textcolor{gray}{84.92}  & \textcolor{gray}{86.93}  & \textcolor{gray}{91.86}  & \textcolor{gray}{87.78}  & \textcolor{gray}{82.29} \\
\bottomrule
\end{tabular}%
}
    \vspace{-0.8em}
\end{table*}

\paragraph{Findings.}
Within the general-MLLM group, AnomalyClaw + GPT-5.5 reaches $79.15\%$ average accuracy --- the best of any general MLLM in this comparison (vs Qwen2.5-VL-72B $76.96$, GPT-5.5 Direct $76.20$) --- and recovers $+2.95$ pp over its own Direct on the same backbone, with the largest wins on tasks that require reference-driven contrast (Defect Analysis $+7.89$, Defect Localization $+5.39$, Object Analysis $+5.02$). AD-specialist methods (gray rows) match or exceed AnomalyClaw on a few subtasks but at the cost of cross-domain generalization (\S\ref{ssec:three_regimes}; AD-Copilot drops $-3.8$ pp on CrossDomainVAD-12 vs its base). MMAD is a cross-task generalization check, not the headline benchmark; full per-subtask numbers and AD-letter-accuracy decomposition are in Appendix~\ref{app:mmad}.

\subsection{Verbalized self-evolution from branch disagreement}
\label{ssec:osr_results}

\paragraph{Setup and headline.} On top of the main agent (Table~\ref{tab:osr_main}, fixed $\alpha{=}0.5$) we compare \textbf{\textsc{OSR}} against three baselines: \textsc{Passive} (no rules; equals the main-table AnomalyClaw); \textsc{$\alpha$-tune} (weight tuning, $K{=}10$ oracle labels/domain spent fitting a per-domain blend $\alpha_d$); and \textsc{+Cluster} (rule learning, same $K{=}10$ budget spent writing rules from a balanced FN/FP batch). \textsc{OSR} learns rules online from branch disagreement with zero oracle labels. We run on Qwen3.5-VL-27B (most \textsc{Passive} headroom) and GPT-5.5 (strongest). \textsc{OSR} directionally lifts both: Qwen3.5-VL-27B reaches $\mathbf{0.7689}$ ($+2.09$pp, $95\%$ CI $[-1.56, +5.60]$, $P(\Delta_{\text{boot}}{>}0){=}0.87$) and GPT-5.5 reaches $\mathbf{0.8225}$ ($+0.90$\, pp); both intervals cross zero, so we treat \textsc{OSR} as an exploratory extension rather than a statistically established improvement. The rule store starts empty and accumulates $56$ rules on Qwen3.5 and $71$ on GPT-5.5, every one triggered by item-level disagreement $|s_{D}{-}s_{R}|{>}0.30$ --- no oracle labels. Per-domain gains concentrate on D6 SDNET cracks ($+19.7$\,pp), D2 VisA ($+6.7$), D10 BMAD-Liver ($+3.8$), D3 MVTec-LOCO ($+2.8$), and D9 BraTS ($+1.9$); full breakdown in Appendix~\ref{app:osr_per_domain}.

\begin{table}[t]
\centering
\caption{\textbf{Verbalized self-evolution.} Macro AUROC on CrossDomainVAD-12 test. $\Delta$ is the macro difference vs.\ \textsc{Passive} and bracketed CIs are stratified paired-bootstrap $95\%$ intervals with $P{=}P(\Delta_{\text{boot}}{>}0)$. Per-domain breakdown in Appendix~\ref{app:osr_per_domain}.}
\label{tab:osr_main}
\small
\setlength{\tabcolsep}{4pt}
\resizebox{\linewidth}{!}{%
\begin{tabular}{llccccc}
\toprule
Backbone & Mode & Oracle/dom & Mechanism & Macro AUROC & $\Delta$ vs Passive [95\% CI], $P$ \\
\midrule
\multirow{5}{*}{Qwen3.5-VL-27B}
& \textsc{Passive}             & $0$  & ---           & $0.7480$ & --- (baseline) \\
& \textsc{+Anchor}             & $0$  & no learning   & $0.7731$ & $+2.51$ [$-0.77, +5.69$], $P{=}0.92$ \\
& \textsc{$\alpha$-tune}       & $10$ & weight tuning & $0.7459$ & $-0.21$ [$-0.99, +0.59$], $P{=}0.29$ \\
& \textsc{+Cluster}            & $10$ & rule learning & $0.7679$ & $+1.99$ [$-1.21, +5.21$], $P{=}0.89$ \\
& \textbf{\textsc{OSR}} (ours) & $\mathbf{0}$ & \textbf{rule learning} & $\mathbf{0.7689}$ & $\mathbf{+2.09}$ [$\mathbf{-1.56, +5.60}$], $\mathbf{P{=}0.87}$ \\
\midrule
\multirow{4}{*}{GPT-5.5}
& \textsc{Passive}             & $0$  & ---           & $0.8135$ & --- (baseline) \\
& \textsc{$\alpha$-tune}       & $10$ & weight tuning & $0.8200$ & $+0.65$ [$-0.11, +1.46$], $P{=}0.95$ \\
& \textsc{+Cluster}            & $10$ & rule learning & $0.8200$ & $+0.65$ [$-1.90, +3.64$], $P{=}0.71$ \\
& \textbf{\textsc{OSR}} (ours) & $\mathbf{0}$ & \textbf{rule learning} & $\mathbf{0.8225}$ & $\mathbf{+0.90}$ [$\mathbf{-2.14, +4.02}$], $\mathbf{P{=}0.70}$ \\
\bottomrule
\end{tabular}%
}
    \vspace{-0.8em}
\end{table}

\paragraph{Rule lever vs weight lever at the same label budget.} \textsc{$\alpha$-tune} ($K{=}10$ oracle labels per domain spent fitting a per-domain blend $\alpha_d$) is flat on Qwen3.5 ($-0.21$\,pp, $P{=}0.29$) and marginally positive on GPT-5.5 ($+0.65$\,pp); \textsc{+Cluster} spends the same $K{=}10$ budget writing rules and reaches $+1.99$\,pp on Qwen3.5 and $+0.65$\,pp on GPT-5.5. Zero-oracle \textsc{OSR} matches \textsc{+Cluster} on Qwen3.5 ($+2.09$ vs $+1.99$\,pp) and slightly exceeds it on GPT-5.5 ($+0.90$ vs $+0.65$\,pp), spending no oracle labels at all. The gain magnitude tracks \textsc{Passive} headroom (GPT-5.5 starts $+6.6$\,pp above Qwen3.5, leaving less room to lift), and the rule lever's edge over the weight lever is consistent with the disagreement set covering items where the agent is \emph{internally unsure} --- a more informative regime for verbalised rules than \textsc{+Cluster}'s narrow FN/FP error slice.

\paragraph{Limitations.} The mechanism requires sufficient internal disagreement to bootstrap: domains where \textsc{Passive}'s two branches happen to agree ($\ge\!90\%$ of items at $|s_{D}{-}s_{R}|{\le}0.30$, e.g.\ D2 VisA and D9 BraTS on Qwen3.5) never fill the $N{=}10$ queue and stay at the \textsc{Passive} baseline. \textsc{OSR} also inherits the Direct-vs-refutation asymmetry: on D8 DermaMNIST, the reflector wrote rules favoring the weaker branch, costing $-5.0$ pp. A confidence gate that requires multi-batch rule agreement before deployment is a natural extension.

\subsection{Behavioural diagnostics}
\label{ssec:behavior}

\begin{figure}[t]
    \centering
    \includegraphics[width=1.0\linewidth]{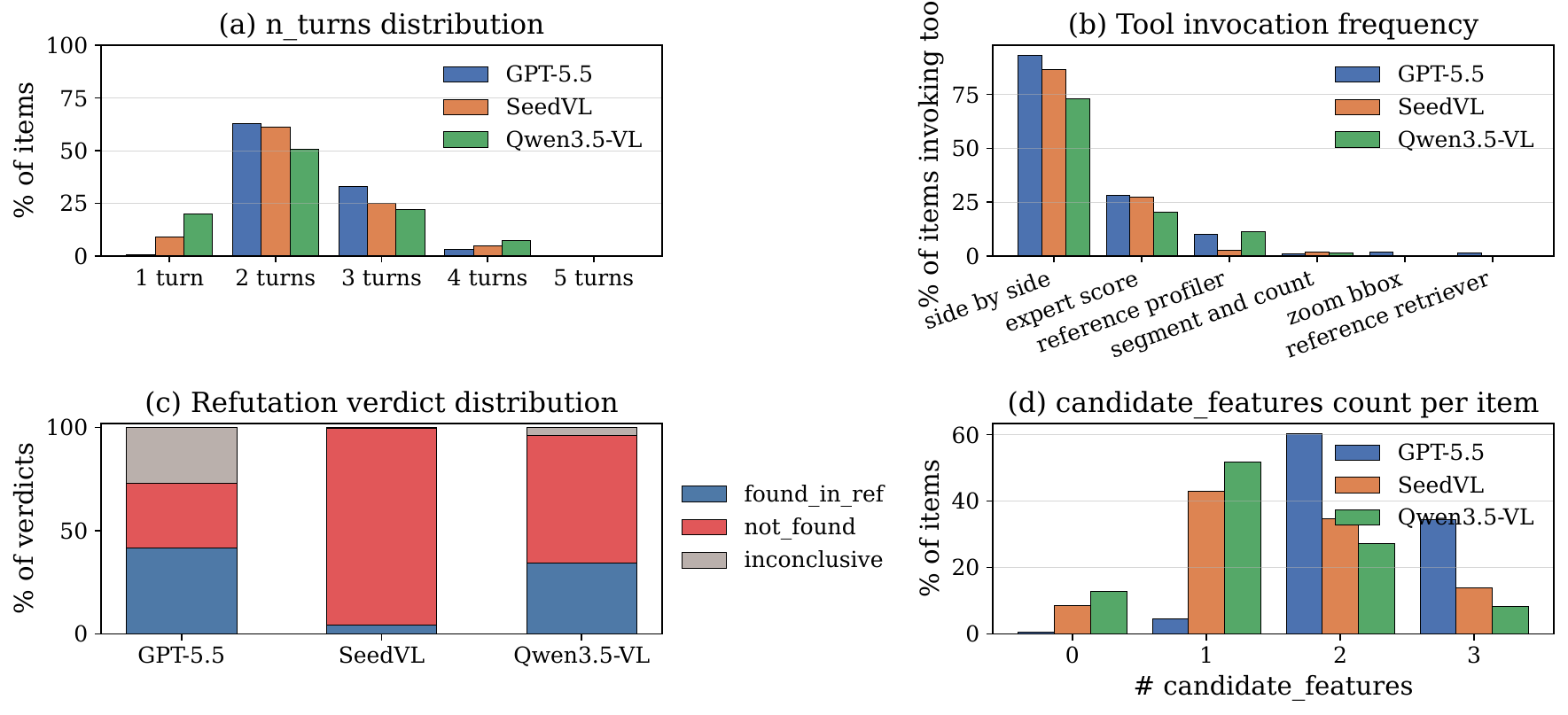}
    \vspace{-0.8em}
    \caption{\textbf{Refutation-trajectory behaviour on CrossDomainVAD-12 test, three VLM backbones} (\textbf{Qwen3.5-VL} denotes Qwen3.5-VL-27B). (a) Per-item turn distribution. (b) Tool invocation frequency, top-$6$ most-invoked tools out of the full $13$-tool catalog. (c) Refutation verdict distribution. (d) \texttt{candidate\_features} count per item.}
    \label{fig:agent_behavior}
    \vspace{-1em}
\end{figure}

We instrument every agent run to record the per-item turn count, the tools invoked, the refutation verdicts, and the candidate-feature list (Figure~\ref{fig:agent_behavior}), and three patterns hold across backbones. Reasoning depth tracks backbone strength: stronger backbones reason longer per item, while weaker ones finalize at turn $1$ far more often. The effective tool catalog is small but distributed, with $4$--$5$ tools each invoked on $\ge 1\%$ of items per backbone and the long tail kept non-zero by per-tool applicability clauses. The refutation-verdict distribution varies sharply between backbones, from a balanced verdict mix on GPT-5.5 to a near-monolithic \texttt{not\_found} preference on Seed2.0-lite. All three are runtime-observable without any test labels and serve as deployment-triage signals for whether a backbone can carry the refutation trajectory in a target domain. Appendix~\ref{app:behavior} reports the per-backbone numbers; further analyses cover the per-branch and blend-weight ablations (Appendices~\ref{app:branch_ablation},~\ref{app:alpha_sensitivity}), leave-one-domain-out and full paired-bootstrap robustness (Appendices~\ref{app:loo},~\ref{app:bootstrap}), the rank-granularity mechanism (Appendix~\ref{app:mechanism}), per-tool invocation rates and compute budget (Appendix~\ref{app:compute_budget}), tool provenance and leakage controls (Appendix~\ref{app:tool_provenance}), beyond-AUROC deployment metrics (Appendix~\ref{app:beyond_auroc}), and the failed-variant design history (Appendix~\ref{app:failed_variants}).

\section{Conclusion}
\label{sec:conclusion}

We frame visual anomaly detection as a compositional perception task and operationalize it as a multi-turn prompt-time agent on a frozen vision-language model. Its core inductive bias is \emph{refutation} rather than evidence accumulation, since anomalousness is defined by what cannot be matched against the references. Built purely at prompt time with no parameter updates, the same agent recipe transfers across vision-language backbones and to new domains. Cross-domain VAD also has no oracle, so we use the agent's own internal branch disagreement as a substitute supervisory signal, growing a verbalized rulebook that improves the agent in place. More broadly, backbone-agnostic agents and label-free internal-feedback loops form a complementary pair that lets a single recipe ride future VLM improvements while deployments improve themselves in unlabelled regimes. We hope CrossDomainVAD-12 and the released agent will seed follow-on work on training-free perception agents that transfer across backbones, domains, and tasks.

\bibliography{references}
\bibliographystyle{unsrtnat}

\newpage
\appendix
\section{Benchmark Specification}
\label{app:benchmark}

Table~\ref{tab:benchmark_detail} provides per-domain details for CrossDomainVAD-12. All splits, item IDs, reference mappings, and domain descriptors will be released as a JSON specification file alongside the code.

\begin{table*}[t]
\centering
\caption{CrossDomainVAD-12 per-domain specification.}
\label{tab:benchmark_detail}
\footnotesize
\setlength{\tabcolsep}{4pt}
\resizebox{\linewidth}{!}{%
\begin{tabular}{cllp{0.32\linewidth}rl}
\toprule
ID & Domain & Source & Anomaly definition & Refs/item & License \\
\midrule
D1  & Industrial  & MVTec-AD & Surface defect, damage, contamination & $\le 10$ & CC BY-NC-SA 4.0 \\
D2  & Complex industrial  & VisA & Surface defect, chips, scratches, missing solder & $\le 10$ & CC BY 4.0 \\
D3  & Logical  & MVTec-LOCO & Wrong count / arrangement / component type & $\le 10$ & CC BY-NC-SA 4.0 \\
D4  & 3D product (RGB render) & Real3D-AD & Geometric defect (bulge, sink, hole, asymmetry) & $\le 10$ & Academic use only$^{\dagger}$ \\
D5  & Retail  & GoodsAD & Torn label, deformation, missing component & $\le 10$ & Academic use only$^{\dagger}$ \\
D6  & Infrastructure  & SDNET2018 & Cracks, spalling, structural damage & $\le 10$ & CC BY-SA 4.0 \\
D7  & Remote-sensing change  & LEVIR-CD+ & Building-level change between temporal pair & 1--2 & Academic use only$^{\dagger}$ \\
D8  & Dermoscopy  & DermaMNIST & Malignant melanoma & $\le 10$ & CC BY 4.0 \\
D9  & Brain MRI  & BraTS2021 & Glioma, mass effect, oedema & $\le 10$ & CC BY 4.0 (DUA) \\
D10 & Liver CT  & BMAD-Liver & Focal lesion, tumour, cyst & $\le 10$ & Academic use only$^{\dagger}$ \\
D11 & GI endoscopy  & HyperKvasir & Polyp, ulcer, inflammation & $\le 10$ & CC BY 4.0 \\
D12 & Road safety  & BDD100K + RoadAnomaly21 & Unexpected object on roadway & 2 & BDD100K license$^{\ddagger}$ \\
\bottomrule
\end{tabular}%
}
\end{table*}

\paragraph{License notes.}
$^{\dagger}$ Real3D-AD, GoodsAD, LEVIR-CD+, BMAD-Liver: distributed for academic/non-commercial research use under each source paper's stated terms; we redistribute only the per-image identifiers and references to original URLs, not the underlying images. $^{\ddagger}$ BDD100K is released under the Berkeley DeepDrive license (research / non-commercial); the RoadAnomaly21 component is academic-research only. BraTS2021 (D9) requires a Data Use Agreement available from the BraTS organizers. Our released benchmark manifest contains only metadata and per-image hashes; users must obtain the underlying datasets from their original distributors and follow each source's license. The CrossDomainVAD-12 manifest itself will be released under MIT.

\paragraph{Reference selection.}
References are normal images from the same category/scene, selected deterministically by metadata match, then fixed random seed. No reference appears as a query. Patient-level (D8--D11) and event-level (D7) deduplication prevents leakage. For D4 (Real3D-AD), the reference pool is rendered views of the same 3D template object from multiple viewpoints.

\paragraph{Split protocol.}
Per domain: 20 calibration, 40 dev, 120 test (balanced 50/50, except D7 LEVIR-CD+, which contributes 98 test items after allocating the 20/40 calibration/dev splits from the 158 source bi-temporal pairs available). All splits frozen before experiments. Total test $n{=}1{,}418$.

\paragraph{Task-anchored domain descriptors.}
The task-anchored descriptor $d$ is a single sentence inserted into every VLM prompt for the \emph{legacy} Direct baseline of Finding~1;  the ensemble's Direct branch (Table~\ref{tab:main_results}) runs \emph{without} these descriptors, matching the refutation agent's descriptor-free task preamble. The twelve descriptors used in CrossDomainVAD-12 are:
\begin{itemize}\small\setlength\itemsep{1pt}
    \item \textbf{D1 Industrial (MVTec-AD):} Normal = a manufactured object matching the reference in geometry, texture, and color. Anomaly = surface defect, damage, contamination, or missing/extra component.
    \item \textbf{D2 Complex industrial (VisA):} Normal = a manufactured object matching the reference. Anomaly = surface defect, chips, scratches, missing solder, or misalignment.
    \item \textbf{D3 Logical (MVTec-LOCO):} Normal = correct count and spatial arrangement of components. Anomaly = wrong count, wrong position, or wrong component type.
    \item \textbf{D4 3D product (Real3D-AD):} Normal = geometrically intact product matching the reference shape under rendered point-cloud views. Anomaly = geometric defect (bulge, sink, hole, contamination, asymmetry).
    \item \textbf{D5 Retail (GoodsAD):} Normal = a packaged supermarket product matching the reference in label, shape, and surface. Anomaly = torn label, deformation, missing component, or printing defect.
    \item \textbf{D6 Infrastructure (SDNET2018):} Normal = intact concrete surface. Anomaly = crack, spalling, or structural damage.
    \item \textbf{D7 Remote-sensing change (LEVIR-CD+):} Normal = no building change between reference and query. Anomaly = building-level change, including new construction, demolition, or expansion. \emph{Planned urbanization IS the anomaly}.
    \item \textbf{D8 Dermoscopy (DermaMNIST):} Normal = benign melanocytic skin lesion. Anomaly = malignant melanoma.
    \item \textbf{D9 Brain MRI (BraTS2021):} Normal = healthy brain MRI slice. Anomaly = glioma, mass effect, or peritumoural edema.
    \item \textbf{D10 Liver CT (BMAD-Liver):} Normal = healthy liver CT slice. Anomaly = focal lesion, tumor, or cyst.
    \item \textbf{D11 GI endoscopy (HyperKvasir):} Normal = clean gastrointestinal mucosa. Anomaly = polyp, ulcer, or inflammation.
    \item \textbf{D12 Road safety (BDD100K + RoadAnomaly21):} Normal = empty driveable roadway (possibly with expected vehicles and markings). Anomaly = unexpected object on the roadway.
\end{itemize}

A corresponding generic baseline descriptor used in Finding~1 reads simply: ``\emph{Identify whether the query image is anomalous relative to the reference(s).}'' The ensemble's Direct branch uses this generic prompt by default.

\section{Legacy Score Extraction Rules (pre-refutation)}
\label{app:scoring}

\emph{This section documents the score-extraction rules used by an earlier router-based agent variant (Routes A/B/C/D). It is retained as a reproducibility artifact and does not describe the current refutation agent in the main paper; for the current scoring rules see \S\ref{ssec:logit_direct} and Appendix~\ref{app:refutation}.}

All VLM variants in the legacy router system output structured JSON. The anomaly score $s \in [0, 1]$:

\paragraph{Direct (legacy).}
Output: \texttt{\{image\_label, confidence, anomaly\_type\}}.
$s = c$ if label $=$ ``anomalous'', $s = 1{-}c$ if ``normal'', where $c$ is reported confidence. Parse failure rate: $<1\%$ across all backbones.

\paragraph{AnomalyClaw interpret (legacy).}
Same JSON schema and scoring. Final score: interpret output on escalated items, $s_0$ on committed items.

\paragraph{Route B (legacy).}
$s = 0.3 \cdot s_0 + 0.7 \cdot \sigma(\text{expert\_score})$, where $\sigma$ normalises relative to calibration median.

\section{Failed agent variants}
\label{app:failed_variants}
\label{app:ablation}

Table~\ref{tab:ablation_appendix} reports the six alternative agent designs we tested on GPT-5.5 calibration. All were fixed and tuned on the same 220-item calibration split as AnomalyClaw. Failure analysis per variant:

\begin{table}[h]
\centering
\caption{Ablation of agent variants (GPT-5.5, calibration macro AUROC). All use the same descriptor-enhanced Direct as the base.}
\label{tab:ablation_appendix}
\small
\begin{tabular}{lcc}
\toprule
Variant & Macro & $\Delta$ vs Direct \\
\midrule
Direct (baseline)     & 0.785 & --- \\
Normal-First          & 0.761 & $-$2.4 \\
Self-Refine           & 0.726 & $-$5.9 \\
Debate (naive score)  & 0.713 & $-$7.2 \\
Debate (gated score)  & 0.744 & $-$4.1 \\
Evidence-grounded prop.  & 0.683 & $-$10.2 \\
Third-call arbiter       & 0.694 & $-$9.1 \\
\midrule
\textbf{AnomalyClaw}     & \textbf{0.837} & \textbf{$+$5.2} \\
\bottomrule
\end{tabular}
\end{table}

\begin{itemize}
    \item \textbf{Normal-First}: structured prompting forces the VLM to enumerate ``normal'' claims, introducing false positives. On Qwen3.5, the reasoning chain consumes the token budget before emitting JSON.
    \item \textbf{Self-Refine}: a second non-adversarial pass adds no new information; it is the weakest variant, confirming that more calls $\neq$ better.
    \item \textbf{Debate}: a refuter rationalises confident correct predictions. Even with a confidence-gated aggregation rule, debate never beats Direct.
    \item \textbf{Evidence injection}: feeding DINOv2 patch distances into the proposer prompt distracts the VLM ($-10$ pp).
    \item \textbf{Arbiter}: a third VLM call on contested items fires on 66\% of items (too aggressive) and is net negative.
\end{itemize}

These ablations were run on GPT-5.5 calibration because they represent disqualified designs rather than competitive baselines; we did not re-run them on SeedVL or Qwen3.5 once the failure pattern was clear.

\section{Expert model comparison (legacy taxonomy)}
\label{app:expert}

This expert-model comparison was conducted during the early design-exploration phase on the legacy twelve-domain split with display codes D01--D11. The code-letter mapping onto the taxonomy of Table~\ref{tab:benchmark_detail} is: D01$\to$D1 MVTec-AD, D02$\to$D5 GoodsAD, D03$\to$D6 SDNET, D04$\to$D8 Derma, D05$\to$D9 BraTS, D06$\to$D10 Liver, D07$\to$D11 Kvasir, D08$\to$D7 LEVIR, D09$\to$D12 road, D10$\to$D3 LOCO, D11$\to$D2 VisA. The headline ensemble (Table~\ref{tab:main_results}) does not depend on these experts.

We benchmarked three expert models on the calibration split (Table~\ref{tab:expert_appendix}). SubspaceAD dominates overall, reaching 1.000 AUROC on D11 VisA calibration, where all VLMs score $\leq 0.85$. AnomalyVFM (zero-shot, DINOv2 + LoRA) is uniformly weaker (macro 0.599) and was excluded from the final system. DINOv2-PatchNN complements SubspaceAD on D03 infrastructure (0.80 vs 0.50). The oracle that selects the best expert per domain reaches macro 0.898, indicating 7 pp headroom for multi-expert routing as future work.

\begin{table}[h]
\centering
\caption{Expert model comparison (calibration, macro AUROC).}
\label{tab:expert_appendix}
\small
\begin{tabular}{lcc}
\toprule
Expert & Macro & Strongest domain \\
\midrule
DINOv2-PatchNN & 0.640 & D09 road (1.00) \\
\textbf{SubspaceAD} & \textbf{0.756} & D11 VisA (1.00) \\
AnomalyVFM & 0.599 & D03 infra (0.72) \\
\midrule
Oracle (max/domain) & 0.898 & --- \\
\bottomrule
\end{tabular}
\end{table}

\section{Descriptor ablation (legacy taxonomy)}
\label{app:descriptor}

This ablation supports Finding~1 and uses the earlier D01--D11 display codes on the earlier twelve-domain split; the code mapping to the taxonomy is listed at the top of Appendix~\ref{app:expert}. The ensemble's Direct branch in Table~\ref{tab:main_results} runs descriptor-free by design (\S\ref{sec:method}), so the ablation is an orthogonal finding about prompt engineering on the Direct baseline rather than a component of the ensemble.

Table~\ref{tab:descriptor_ablation} shows the per-domain effect of replacing the generic descriptor (``Identify whether the query image is anomalous relative to the reference(s)'') with the task-anchored descriptor (Appendix~\ref{app:benchmark}) on the test split for all three backbones. The macro gain is positive and statistically significant for every backbone (paired bootstrap with 1{,}000 resamples, per-domain stratification, $n{=}1{,}298$): GPT-5.5 $+6.4$ pp [$+4.4, +8.4$], SeedVL $+4.1$ pp [$+1.6, +6.4$], Qwen3.5 $+3.2$ pp [$+1.0, +5.5$]. The three backbones agree on the qualitative direction but disagree on which domains benefit most: GPT-5.5 is dominated by D06 Liver CT and D08 Change ($+31$/$+33$ pp), SeedVL gains most on D08 Change and D09 Road ($+18$/$+14$ pp), Qwen3.5 gains most on D06 Liver and D04 Dermoscopy ($+14$/$+8$ pp). A handful of small per-domain regressions appear on each backbone (e.g.\ D03 Infrastructure on all three, D09 Road on Qwen3.5); these are VLM-strong domains where the more specific prompt trims some true-positive confidence.

\begin{table}[h]
\centering
\caption{Descriptor ablation on the test split: per-domain AUROC (Generic vs Task-anchored, $\Delta$ in pp). Bold = largest per-domain gain per backbone.}
\label{tab:descriptor_ablation}
\footnotesize
\setlength{\tabcolsep}{4pt}
\begin{tabular}{lccc|ccc|ccc}
\toprule
& \multicolumn{3}{c|}{GPT-5.5} & \multicolumn{3}{c|}{SeedVL} & \multicolumn{3}{c}{Qwen3.5} \\
Domain & Gen. & Task & $\Delta$ & Gen. & Task & $\Delta$ & Gen. & Task & $\Delta$ \\
\midrule
D01 Industrial      & 0.951 & 0.962 & $+$1.0  & 0.848 & 0.874 & $+$2.6 & 0.912 & 0.903 & $-$0.8 \\
D02 Retail          & 0.752 & 0.774 & $+$2.2  & 0.864 & 0.863 & $-$0.1 & 0.632 & 0.672 & $+$4.0 \\
D03 Infrastructure  & 0.664 & 0.623 & $-$4.1  & 0.745 & 0.660 & $-$8.6 & 0.702 & 0.712 & $+$1.0 \\
D04 Dermoscopy      & 0.769 & 0.796 & $+$2.7  & 0.708 & 0.760 & $+$5.2 & 0.680 & 0.762 & \textbf{$+$8.2} \\
D05 Brain MRI       & 0.915 & 0.934 & $+$1.9  & 0.736 & 0.864 & \textbf{$+$12.8} & 0.866 & 0.849 & $-$1.8 \\
D06 Liver CT        & 0.433 & 0.745 & \textbf{$+$31.2} & 0.529 & 0.492 & $-$3.7 & 0.544 & 0.684 & \textbf{$+$14.0} \\
D07 GI Endoscopy    & 0.903 & 0.890 & $-$1.3  & 0.881 & 0.876 & $-$0.4 & 0.874 & 0.918 & $+$4.4 \\
D08 Change Det.     & 0.496 & 0.827 & \textbf{$+$33.2} & 0.643 & 0.822 & \textbf{$+$17.9} & 0.767 & 0.828 & $+$6.1 \\
D09 Road Obstacle   & 0.954 & 0.968 & $+$1.4  & 0.793 & 0.936 & \textbf{$+$14.3} & 0.970 & 0.911 & $-$6.0 \\
D10 Logical         & 0.660 & 0.683 & $+$2.3  & 0.678 & 0.651 & $-$2.7 & 0.641 & 0.676 & $+$3.5 \\
D11 VisA            & 0.873 & 0.878 & $+$0.5  & 0.805 & 0.878 & $+$7.3 & 0.776 & 0.800 & $+$2.4 \\
\midrule
Macro               & \textbf{0.761} & \textbf{0.825} & \textbf{$+$6.4} &
                      \textbf{0.748} & \textbf{0.789} & \textbf{$+$4.1} &
                      \textbf{0.760} & \textbf{0.792} & \textbf{$+$3.2} \\
95\% CI ($\Delta$)  & \multicolumn{3}{c|}{$[+4.4, +8.4]$} &
                      \multicolumn{3}{c|}{$[+1.6, +6.4]$} &
                      \multicolumn{3}{c}{$[+1.0, +5.5]$} \\
\bottomrule
\end{tabular}
\end{table}

\section{Per-backbone macro and per-domain AUROC}
\label{app:perdomain}

\begin{figure}[h]
\centering
\includegraphics[width=0.92\linewidth]{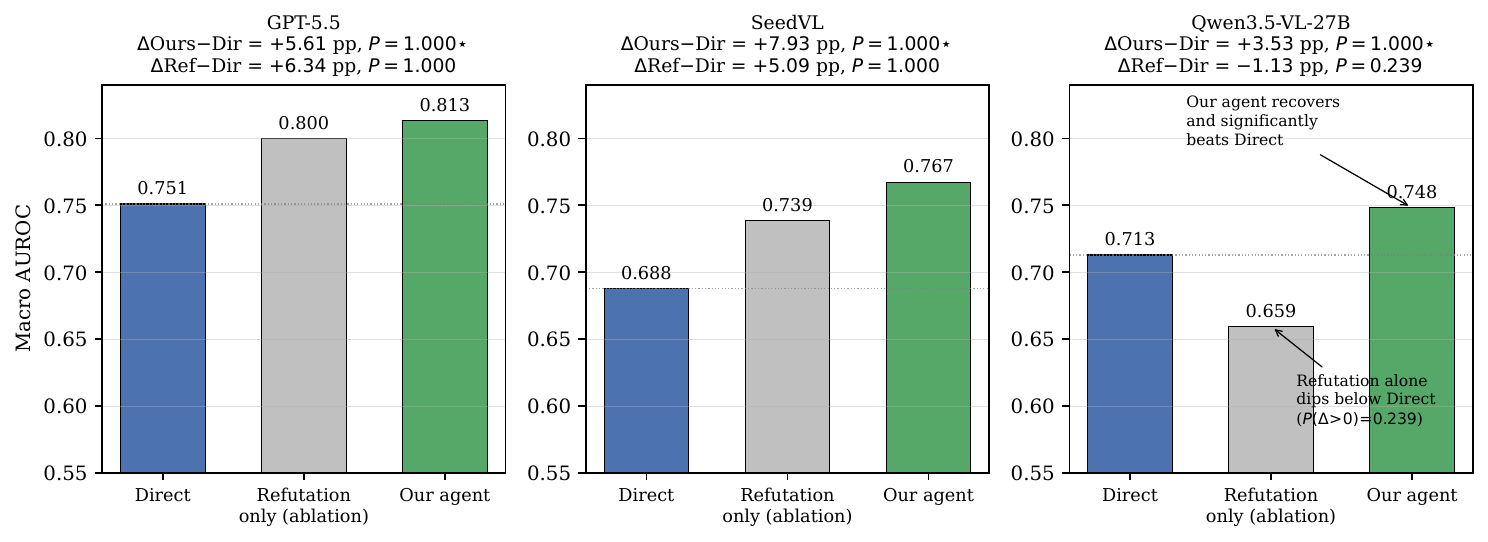}
\caption{\textbf{Per-backbone macro AUROC: the parallel-Direct branch is the safety net.} Macro AUROC on \textsc{CrossDomainVAD-12} test for three VLM backbones. Each panel shows Direct (single-pass VLM baseline), Refutation only (ablation: our agent with the parallel-Direct branch removed), and our agent (the two branches averaged inside a single per-item invocation). On GPT-5.5 and SeedVL the refutation trajectory is individually strong; on Qwen3.5-VL-27B it dips below Direct in mean ($\Delta{=}-1.13$ pp, $P{=}0.239$ --- not statistically separable from Direct, but with substantial probability of regressing) and shows dramatic per-domain failures (D11 GI endoscopy $-24.4$ pp, D7 LEVIR $-10.1$ pp; see Figure~\ref{fig:per_domain_appendix} right panel). Our agent restores the macro to a $+3.52$ pp gain over JSON Direct under apples-to-apples JSON-fusion ($95\%$ CI $[+1.93,+5.11]$; $0/1{,}000$ bootstrap resamples $\le 0$, $P(\Delta{>}0){>}0.999$). All three backbones are significant at $95\%$.}
\label{fig:hero_appendix}
\end{figure}

\begin{figure}[h]
\centering
\includegraphics[width=\linewidth]{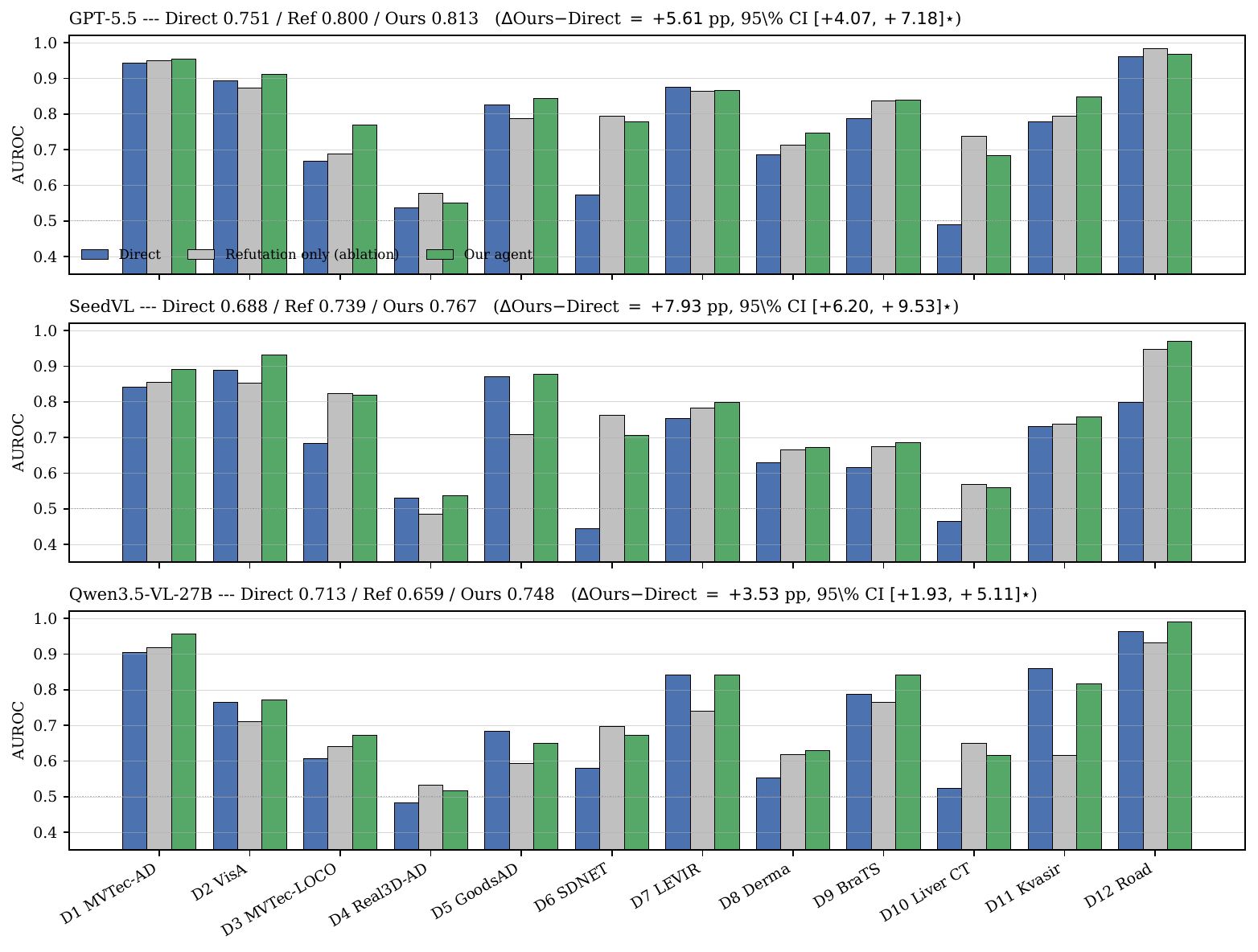}
\caption{\textbf{Per-domain AUROC on CrossDomainVAD-12 test}, one panel per VLM backbone. Three bars per domain: Direct (blue) / refutation only (gray, ablation: our agent without the parallel-Direct branch) / our agent (green). Grey dotted line marks the $0.5$ random-guessing floor. Macro numbers and the stratified paired bootstrap $\Delta$(Ours$-$Direct) with $95\%$ CI are reported in each panel title. Our agent's complementary-error gain concentrates on infrastructure (D6), liver CT (D10), and road (D12) on SeedVL; on GPT-5.5 the gain is uniform across industrial and medical families; on Qwen3.5-VL-27B our agent compensates for the refutation trajectory's per-domain regressions on D11 (Kvasir, $-24.4$ pp) and D7 (LEVIR, $-10.1$ pp) by relying on Direct.}
\label{fig:per_domain_appendix}
\end{figure}

Figure~\ref{fig:hero_appendix} reports the per-backbone macro story: Direct, refutation-only ablation, and our agent on each of GPT-5.5, SeedVL, and Qwen3.5-VL-27B. Figure~\ref{fig:per_domain_appendix} expands this to the per-domain breakdown on all twelve domains for each backbone. Table~\ref{tab:main_results} contains the exact macro numbers and bootstrap CIs; the figures visualize where our agent's complementary-error advantage concentrates and where the refutation trajectory's per-backbone behavior diverges.

\begin{table}[h]
\centering
\caption{Per-backbone macro AUROC on CrossDomainVAD-12 test, $\ge 90$ valid items per domain after error filtering: Direct (single-pass baseline), refutation-only (ablation: our agent with the parallel-Direct branch removed), and our agent. Headline numbers match Table~\ref{tab:main_results}; full per-domain breakdown in Figure~\ref{fig:per_domain_appendix}. All comparisons use the $1{,}000$-resample stratified paired bootstrap with per-domain stratification.}
\label{tab:per_domain}
\small
\begin{tabular}{lcccccc}
\toprule
Backbone & Direct & Refutation only & Ours & $\Delta_{\text{Ours}-\text{D}}$ & 95\% CI & $P(\Delta{>}0)$ \\
\midrule
GPT-5.5  ($n{=}1{,}418$) & 0.751 & 0.800 & \textbf{0.814} & $+6.23$ & $[+4.84, +7.63]$ & $>0.999$ \\
SeedVL   ($n{=}1{,}418$) & 0.688 & 0.739 & \textbf{0.767} & $+7.93$ & $[+6.20, +9.53]$ & $>0.999$ \\
Qwen3.5  ($n{=}1{,}418$) & 0.713 & 0.701 & \textbf{0.748} & $+3.52$ & $[+1.93, +5.11]$ & $>0.999$ \\
\bottomrule
\end{tabular}
\end{table}

\section{Per-branch ablation: refutation alone vs. direct alone}
\label{app:branch_ablation}

To make the role of each internal branch explicit, we report a controlled ablation that removes the Direct branch and keeps only the multi-turn refutation trajectory. Macro AUROC of the refutation-only ablation vs Direct (paired bootstrap, $1{,}000$ resamples, per-domain stratification): GPT-5.5 $+6.34$ pp, $95\%$ CI $[+3.90, +8.82]$, $0/1{,}000$ resamples $\le 0$ ($P(\Delta{>}0){>}0.999$); Seed2.0-lite $+5.09$ pp, $[+2.48, +7.66]$, $0/1{,}000$ ($P{>}0.999$); Qwen3.5-VL-27B $-1.13$ pp, $[-4.01, +1.76]$, $P(\Delta{>}0){=}0.239$. Two regimes follow.

\paragraph{Productive-refutation regime (GPT-5.5, Seed2.0-lite).}
Both internal branches individually beat Direct. On GPT-5.5, the refutation alone already beats Direct by $+6.34$ pp; AnomalyClaw lands at $+6.23$ pp --- slightly below the refutation-only ablation in mean but with a tighter CI ($[+4.84,+7.63]$) because the parallel-Direct branch reduces variance. On Seed2.0-lite, both branches are productive, and the agent reaches $+7.93$ pp over Direct (CI $[+6.20,+9.53]$), exceeding either branch alone.

\paragraph{Failure-mode regime (Qwen3.5-VL-27B).}
At the macro level, the refutation-only ablation is statistically indistinguishable from Direct ($-1.13$ pp, CI $[-4.01,+1.76]$, $P{=}0.239$): mean negative, but the lower CI bound at $-4.01$ pp says the refutation could plausibly regress Direct by several points. The per-domain picture is sharper: on D11 HyperKvasir GI endoscopy, the refutation alone scores $0.617$ vs Direct's $0.860$, a $-24.4$ pp single-domain regression; on D7 LEVIR change-detection, the refutation alone scores $0.741$ vs Direct's $0.842$, a $-10.1$ pp regression. The parallel-Direct branch absorbs these per-domain regressions: AnomalyClaw's macro $\Delta$ over JSON Direct on Qwen3.5 is $+3.52$ pp ($[+1.93,+5.11]$, $0/1{,}000$ resamples $\le 0$, $P(\Delta{>}0){>}0.999$) under the apples-to-apples JSON-fusion configuration, significant despite per-domain refutation regressions of up to $24$ pp. The Direct branch is the safety net that allows the agent to be deployed on a new VLM backbone without verifying, ahead of time, that multi-turn refutation outperforms single-pass scoring on every domain.

\section{Blend-weight and logit-Direct sensitivity}
\label{app:alpha_sensitivity}
\label{app:logit_direct_ablation}

The main agent fixes $\alpha{=}0.5$ before test-time evaluation, with
\[
s_{\mathrm{final}}=\alpha s_D + (1-\alpha)s_R.
\]
Table~\ref{tab:alpha_sensitivity} sweeps a small post-hoc grid using the saved Direct and refutation scores. The best grid point differs by backbone, but the fixed midpoint is within $0.5$ pp of the best point on all three backbones; the headline result is therefore not a cherry-picked blend weight.

\begin{table}[h]
\centering
\caption{\textbf{Post-hoc sensitivity to the Direct/refutation blend weight.} $\alpha{=}0$ is refutation-only; $\alpha{=}1$ is Direct-only. The paper uses $\alpha{=}0.5$ fixed a priori.}
\label{tab:alpha_sensitivity}
\small
\begin{tabular}{lccccc|cc}
\toprule
Backbone & $\alpha{=}0$ & $0.25$ & $\mathbf{0.5}$ & $0.75$ & $1.0$ & best $\alpha$ & best--$0.5$ \\
\midrule
Qwen3.5-VL-27B & 0.7011 & 0.7437 & \textbf{0.7480} & 0.7485 & 0.7128 & 0.75 & $+0.05$ pp \\
Seed2.0-lite   & 0.7387 & 0.7719 & \textbf{0.7672} & 0.7505 & 0.6879 & 0.25 & $+0.47$ pp \\
GPT-5.5        & 0.8166 & 0.8180 & \textbf{0.8135} & 0.8040 & 0.7508 & 0.25 & $+0.45$ pp \\
\bottomrule
\end{tabular}
\end{table}

For Qwen3.5-VL-27B, vLLM additionally exposes first-token logprobs, so the Direct branch can be scored as a soft Yes/No probability rather than as a JSON label plus reported confidence. We treat this as an \emph{orthogonal score-extraction optimisation}, not as part of the agent's contribution, and report two cleanly separated effects:
\begin{itemize}\setlength\itemsep{1pt}
\item \textbf{Agent contribution (apples-to-apples).} Holding the Direct form fixed at JSON confidence, AnomalyClaw lifts macro AUROC from $0.7128$ (JSON Direct) to $0.7480$, $+3.52$ pp. This is the configuration reported in Table~\ref{tab:main_results} and uses the same protocol as the GPT-5.5 and Seed2.0-lite rows.
\item \textbf{Score-extraction optimisation (Qwen-only).} Replacing JSON-confidence with logit-as-soft-label in the Direct branch lifts the standalone Direct macro from $0.7128$ to $0.7643$, $+5.15$ pp. This is a property of the readout, not of the agent: it would lift any system that consumes the Direct score, and is unavailable on GPT-5.5 / Seed2.0-lite because their chat-completions APIs do not expose logprobs.
\item \textbf{Composed (deployment).} Fusing the agent with the logit-Direct form gives a final macro of $0.7672$, which can be read either as $+5.42$ pp over JSON Direct (mixing both effects) or as $+0.30$ pp over logit Direct alone. Neither composed number is a clean attribution to the agent in isolation; for that, see the apples-to-apples bullet above.
\end{itemize}
The agent's contribution on Qwen3.5 is therefore $+3.52$ pp, identical in magnitude and CI shape to the appendix branch-ablation table, irrespective of which Direct readout the deployment subsequently uses.

\section{Mechanism: rank granularity, not middle-mass}
\label{app:mechanism}

Why does combining Direct with the refutation score help, even when the refutation alone is individually weaker (Qwen3.5-VL-27B)? An intuitive candidate mechanism is \emph{middle-zone mass}: the refutation's scores fall in the $[0.2, 0.8]$ range more often than Direct's, and this `intermediate confidence' mass carries a signal that Direct's bimodal cluster misses. Controlled transformations below are inconsistent with this candidate; they instead point to \emph{rank granularity} (the number of unique score values the refutation branch emits) as the primary driver of the gain.

\begin{table}[h]
\centering
\caption{Controlled transformations of the refutation-trajectory score on Qwen3.5-VL-27B test ($n{=}1{,}418$, macro AUROC over $12$ domains). Columns: standalone AUROC of the transformed score; combined AUROC of $0.5{\cdot}\text{Direct}+0.5{\cdot}X$ over JSON Direct (in pp); number of unique transformed values; middle-zone mass in $[0.2,0.8]$. The Direct baseline is the JSON-confidence form (macro $0.713$), matching the apples-to-apples agent attribution in Table~\ref{tab:main_results}; the ``Original'' combined gain reads $+3.52$ pp. \textbf{EXT-RANK} is the key row: middle-mass forced to $0\%$ but every rank preserved --- the gain matches the original. \textbf{BIN} collapses ranks from $14$ to $2$ and the gain halves. Middle-mass and rank granularity correlate on raw outputs; the controlled transformations isolate rank granularity as the consistent driver of the gain.}
\label{tab:rank_granularity}
\small
\resizebox{\linewidth}{!}{%
\begin{tabular}{lccccc}
\toprule
Transformation & Standalone AUROC & Combined $\Delta$ over Direct & Unique vals & Middle mass \\
\midrule
Original (no transform)                        & $0.7011$ & $+3.52$ pp & $14$ & $0.3\%$ \\
BIN (median-split $\to\!\{0.05,0.95\}$)        & $0.6220$ & $+2.33$ pp & $2$  & $0\%$ \\
EXT-RANK (rank-preserving, middle $=0$)        & $0.7011$ & $+3.53$ pp & $14$ & $0\%$ \\
AFFINE, $a{=}0.25$ (compress around $0.5$)     & $0.7011$ & $+3.56$ pp & $14$ & $100\%$ \\
AFFINE, $a{=}2.0$ (expand, clip)               & $0.6510$ & $+2.20$ pp & $2$  & $0\%$ \\
\bottomrule
\end{tabular}%
}
\end{table}

Three clean falsifications follow (Table~\ref{tab:rank_granularity}).
(i) \textbf{Middle-mass does not appear to drive the gain.} EXT-RANK monotone-remaps the $14$ unique refutation scores into $[0.05, 0.19] \cup [0.81, 0.95]$, preserving every rank but forcing middle-zone mass to zero; the gain is $+3.53$ pp, numerically identical to the original $+3.52$ pp. If the middle mass carried the signal, EXT-RANK would have collapsed the gain.
(ii) \textbf{Rank granularity tracks the gain consistently.} BIN collapses the $14$ unique values to $2$ via per-split median split, preserving the sign of every prediction but destroying rank resolution; the gain halves to $+2.33$ pp. AFFINE with $a{>}1$ clips values outside $[0,1]$ and silently reduces the effective rank count to $2$, dropping the gain to $+2.20$ pp.
(iii) \textbf{The positive control is clean.} AFFINE with $a{\le}1$ compresses scores around $0.5$ without clipping, preserving every rank; the gain stays at $+3.56$ pp regardless of middle-mass ($100\%$ at $a{=}0.25$).

On this evidence, the mechanism is specifically: \emph{Direct emits a coarse bimodal rank grid ($11$ unique values on Qwen3.5 test); averaging with the refutation trajectory's $14$-value rank grid breaks Direct's ties inside each bimodal cluster with real signal}. Middle-mass and rank granularity correlate on raw VLM outputs because free-form JSON produces many distinct numerals, but the lab decoupling above isolates rank granularity as the primary driver of the observed gain in this controlled setting.

\section{Specialty-aware tool catalog}
\label{app:specialty}

Our agent's refutation trajectory has access to a $13$-tool catalog. Each tool ships with a structured description and a \emph{per-tool applicability clause} that specifies the domain families in which the tool carries native signal and those in which it is uninformative. The applicability clause is presented to the VLM at every turn alongside the tool's input/output schema, so the VLM can choose tools based on a domain-grounded match rather than tool-name salience alone.

The full $13$ tools, with abridged applicability clauses:
\begin{itemize}\setlength\itemsep{0pt}
\item \texttt{tool\_side\_by\_side} --- crop query and references at matching bounding boxes; applicable to all domains.
\item \texttt{tool\_reference\_profiler} --- structured normal-baseline summary of the reference pool; applicable to all domains.
\item \texttt{tool\_expert\_score} --- query a frozen domain-specialist (subspace AD on D1/D5/D6/D3/D7/D9/D10) for a feature-space anomaly score; applicable on aligned-image domains where a strong handcrafted baseline exists.
\item \texttt{tool\_zoom\_bbox} --- crop and rescale a candidate bbox; applicable to small-defect domains (industrial, infrastructure).
\item \texttt{tool\_image\_diff} --- pixel-level diff against an aligned reference; restricted to aligned-domain set $\{$D1, D5, D3$\}$.
\item \texttt{tool\_rotate\_align} --- rigid alignment to canonical pose; restricted to the aligned-domain set.
\item \texttt{tool\_segment\_and\_count} --- foreground segmentation and component count; applicable to logical (D3) and retail (D5) where countable units carry the anomaly definition.
\item \texttt{tool\_patch\_grid} --- $K{\times}K$ patch comparison; applicable to texture/material domains (D1, D2).
\item \texttt{tool\_texture\_fft} --- FFT spectrum diff; applicable to industrial materials (D1, D2, D6).
\item \texttt{tool\_reference\_retriever} --- retrieve nearest-normal references from the reference pool; applicable to all domains.
\item \texttt{tool\_domain\_knowledge} --- retrieve a one-paragraph domain primer; applicable on under-described medical and remote-sensing domains.
\item \texttt{tool\_component\_counter} --- task-specific instance counter (declared but seldom selected by any backbone).
\item \texttt{tool\_visual\_retriever} --- visual-similarity retriever (declared but seldom selected by any backbone).
\end{itemize}
This per-tool applicability presentation is what we refer to in \S\ref{ssec:behavior} when we report that $4$--$5$ tools see $\ge 1\%$ invocation rate per backbone (Figure~\ref{fig:agent_behavior}b): without applicability clauses, refutation trajectories tend to concentrate on the single most generic tool.

\subsection{Compute budget per item and per-tool invocation rates}
\label{app:compute_budget}

Each per-item invocation runs the Direct branch (one VLM call) and the refutation trajectory (one VLM call per turn) concurrently on the same backbone session. Most of the $13$ catalog tools are frozen Python and do not consume a VLM call; two exceptions issue an additional VLM/LLM call inside the tool body (\texttt{reference\_profiler}, \texttt{domain\_knowledge}). The honest accounting is therefore
\[
\text{VLM calls/item} \;=\; \underbrace{1}_{\text{Direct}} \;+\; n_{\text{turns}} \;+\; n_{\text{extra-VLM tools}},
\]
which we report in Table~\ref{tab:compute_budget}; the per-tool item-level invocation rate is in Table~\ref{tab:tool_usage}; the per-domain breakdown is in Table~\ref{tab:budget_per_domain}. All numbers are computed from the same canonical result manifests as Table~\ref{tab:main_results}.

\begin{table}[h]
\centering
\caption{\textbf{Compute budget per item} on CrossDomainVAD-12 test ($n{=}1{,}418$ items per backbone). \emph{Total VLM calls/item} is the sum of the Direct call, the refutation turns, and any tool-internal VLM call (\texttt{reference\_profiler}, \texttt{domain\_knowledge}). Cost ratio is vs.\ a single-pass Direct baseline ($1$ call/item).}
\label{tab:compute_budget}
\small
\setlength{\tabcolsep}{4pt}
\begin{tabular}{lcccccc}
\toprule
 & \multicolumn{2}{c}{Refutation turns} & \multicolumn{3}{c}{\textbf{Total VLM calls / item}} & \\
\cmidrule(lr){2-3}\cmidrule(lr){4-6}
Backbone & mean & max & mean & p90 & max & cost ratio \\
\midrule
GPT-5.5          & $2.01$ & $3$ & $\mathbf{3.07}$ & $4$ & $5$ & $3.07\times$ \\
Seed2.0-lite     & $2.25$ & $4$ & $\mathbf{3.29}$ & $4$ & $6$ & $3.29\times$ \\
Qwen3.5-VL-27B   & $2.26$ & $5$ & $\mathbf{3.42}$ & $5$ & $8$ & $3.42\times$ \\
\bottomrule
\end{tabular}
\end{table}

\begin{table}[h]
\centering
\caption{\textbf{Per-tool invocation rate} (\% of items that invoke the tool at least once during their refutation trajectory). Tools whose body issues a VLM/LLM call (\texttt{reference\_profiler}, \texttt{domain\_knowledge}) are starred. Variants of the same tool emitted under prefixed and non-prefixed names (e.g.\ \texttt{tool\_segment\_and\_count} and \texttt{segment\_and\_count}) are merged.}
\label{tab:tool_usage}
\small
\begin{tabular}{lccc}
\toprule
Tool & GPT-5.5 & Seed2.0-lite & Qwen3.5-VL-27B \\
\midrule
\texttt{tool\_side\_by\_side}        & $76.4\%$ & $83.5\%$ & $86.2\%$ \\
\texttt{tool\_expert\_score}         & $17.2\%$ & $20.2\%$ & $21.3\%$ \\
\texttt{tool\_reference\_profiler}$^*$   & $\phantom{0}4.8\%$  & $11.1\%$ & $14.2\%$ \\
\texttt{tool\_zoom\_bbox}            & $\phantom{0}0.4\%$  & $\phantom{0}0.4\%$  & $\phantom{0}1.0\%$ \\
\texttt{tool\_reference\_retriever}  & $\phantom{0}1.1\%$  & $\phantom{0}0.1\%$  & $\phantom{0}0.4\%$ \\
\texttt{tool\_segment\_and\_count}   & $\phantom{0}0.1\%$  & $\phantom{0}1.3\%$  & $\phantom{0}2.2\%$ \\
\texttt{tool\_image\_diff}           & $\phantom{0}0.1\%$  & $-$    & $-$ \\
\texttt{tool\_domain\_knowledge}$^*$  & $\phantom{0}0.9\%$  & $\phantom{0}0.4\%$  & $\phantom{0}1.1\%$ \\
\texttt{tool\_patch\_grid}           & $-$    & $\phantom{0}0.1\%$  & $\phantom{0}0.2\%$ \\
\texttt{tool\_rotate\_align}, \texttt{tool\_texture\_fft}, others & $<\!0.1\%$ & $<\!0.1\%$ & $<\!0.2\%$ \\
\midrule
\textbf{Effective tools at $\ge\!1\%$} & \textbf{4} & \textbf{5} & \textbf{6} \\
\bottomrule
\end{tabular}
\end{table}

\begin{table}[h]
\centering
\caption{\textbf{Compute budget per domain} (mean VLM calls/item across all $n=1{,}418$ items, one column per backbone). Compute concentrates on domains where Direct is weakest --- D6 SDNET cracks, D9 BraTS brain MRI, D10 BMAD-Liver, D7 LEVIR change --- consistent with the agent spending refutation turns where the single-pass call provides the least signal.}
\label{tab:budget_per_domain}
\small
\setlength{\tabcolsep}{6pt}
\begin{tabular}{lccc}
\toprule
Domain (D-code) & GPT-5.5 & Seed2.0-lite & Qwen3.5-VL-27B \\
\midrule
D1 \;Industrial (MVTec-AD)      & $2.81$ & $2.79$ & $3.49$ \\
D2 \;Complex industrial (VisA)  & $2.67$ & $2.78$ & $2.99$ \\
D3 \;Logical (MVTec-LOCO)       & $2.83$ & $2.87$ & $3.58$ \\
D4 \;3D product (Real3D-AD)     & $3.21$ & $3.45$ & $3.29$ \\
D5 \;Retail (GoodsAD)           & $2.76$ & $2.96$ & $3.08$ \\
D6 \;Infrastructure (SDNET2018) & $2.83$ & $3.08$ & $3.02$ \\
D7 \;Remote-sensing (LEVIR-CD+) & $3.15$ & $3.61$ & $4.13$ \\
D8 \;Dermoscopy (DermaMNIST)    & $3.07$ & $3.12$ & $3.06$ \\
D9 \;Brain MRI (BraTS2021)      & $3.67$ & $3.71$ & $4.07$ \\
D10 Liver CT (BMAD-Liver)       & $3.51$ & $3.94$ & $4.06$ \\
D11 GI endoscopy (HyperKvasir)  & $2.96$ & $3.06$ & $3.01$ \\
D12 Road safety (BDD+RoadAnomaly21) & $2.72$ & $2.77$ & $3.01$ \\
\midrule
\textbf{Macro mean}             & $\mathbf{3.07}$ & $\mathbf{3.29}$ & $\mathbf{3.42}$ \\
\bottomrule
\end{tabular}
\end{table}

Three honest observations. \textbf{(i) Cost.} AnomalyClaw averages $3.07$--$3.42$ VLM calls per item across backbones --- a $3.0$--$3.4\times$ inference cost over a single-pass Direct call for a $+6.23$--$+7.93$ pp macro AUROC gain (Table~\ref{tab:main_results}); the macro gain per extra VLM call is roughly $+1.7$--$+3.5$ pp. Wall-time is lower than the call count would suggest because the Direct branch and the refutation trajectory are issued concurrently on the same backbone session, but \emph{token billing} (which dominates closed-API cost) scales with the call count, not wall-time. \textbf{(ii) Where compute concentrates.} The agent spends most turns on domains where the single-pass Direct is weakest: D6 SDNET cracks, D9 BraTS, D10 BMAD-Liver, and D7 LEVIR remote-sensing change all average $3.5$--$4.1$ calls/item across backbones, while industrial-clean domains (D1 MVTec-AD, D5 GoodsAD, D12 road) finish in $2.7$--$3.1$ calls/item. This per-domain compute distribution is itself a runtime-observable signal of where the refutation step is being exercised, available without any test labels. \textbf{(iii) Backbone effect.} Stronger reasoning backbones commit faster: GPT-5.5 averages $2.01$ refutation turns per item versus $2.26$ for Qwen3.5-VL-27B, and the GPT-5.5 max is $3$ turns versus $5$ for Qwen3.5. The $K{=}5$ turn budget is non-binding on average for all backbones but \emph{is} reached in the long tail on Qwen3.5 (max $5$ turns; max VLM calls $8$/item including a long-tailed extra \texttt{reference\_profiler} call), so the cap is real for the weakest backbone.

\paragraph{Caveats and what is not measured.}
We report VLM call counts, not tokens billed or wall-clock latency. Tokens depend on per-turn prompt length (system prompt, image input, accumulated tool observations, JSON output schema), which we did not log uniformly across backbones. Closed-API cost in dollars is approximately proportional to call count $\times$ per-call tokens, but the per-call token mix is backbone- and turn-specific. Iso-budget comparisons against single-pass Direct variants spending equal token budget (e.g., Direct$@K{=}3$ self-consistency, Direct$+$CoT, Direct$+$reference profiling) are not yet reported and are a known limitation; we expect them to close some but not all of the AnomalyClaw gain, especially on the cross-domain shift cases (D7, D9, D10) where single-pass Direct errors look qualitatively different from low-confidence Direct.

\section{Refutation-trajectory behavioural diagnostics}
\label{app:behavior}

This appendix expands the three behavioral diagnostics summarised in \S\ref{ssec:behavior} and refers throughout to the four-panel Figure~\ref{fig:agent_behavior} in the body.

\paragraph{Reasoning depth tracks backbone strength.}
The $n$-turns distribution in Figure~\ref{fig:agent_behavior}a is a proxy for how much reasoning the refutation trajectory performs before scoring. GPT-5.5 finalises at turn $1$ on $0.7\%$ of items, Seed2.0-lite on $9.0\%$, and Qwen3.5-VL-27B on $19.8\%$. Mean \texttt{candidate\_features} count is $2.29$ on GPT-5.5 versus $1.31$ on Qwen3.5, and $12.7\%$ of Qwen3.5 items enter final-turn scoring with zero candidate features versus $0.6\%$ on GPT-5.5. The trajectory is not broken on Qwen3.5, just shallower, quantitatively consistent with Qwen3.5's per-domain refutation regressions in \S\ref{ssec:three_regimes}. Reasoning depth is runtime-observable without any test labels and is the cleanest predictor of which backbones the refutation agent can carry on its own.

\paragraph{An effective tool catalog is small but distributed.}
Figure~\ref{fig:agent_behavior}b shows that $4$--$5$ tools each see invocation on $\ge 1\%$ of items per backbone. \texttt{tool\_side\_by\_side} is the workhorse at $73$--$93\%$, \texttt{tool\_expert\_score} at $20$--$28\%$, and \texttt{tool\_reference\_profiler} at $2$--$11\%$, with \texttt{tool\_segment\_and\_count} and \texttt{tool\_zoom\_bbox} on GPT-5.5 rounding out the $1$--$2\%$ tier. The remaining $8$ tools see $\le 0.5\%$ usage. The per-tool applicability clauses in Appendix~\ref{app:specialty} keep the long tail non-zero without runaway concentration on a single tool. Per-tool invocation rates with VLM/non-VLM tool annotations are reported in Table~\ref{tab:tool_usage}.

\paragraph{Refutation-verdict distribution is strongly backbone-dependent.}
The protocol asks the VLM to return a verdict per targeted candidate feature, and a balanced refuter should produce mixed verdicts. GPT-5.5 is most balanced at $41\%$ \texttt{found\_in\_ref}, $32\%$ \texttt{not\_found}, and $27\%$ \texttt{inconclusive}. Seed2.0-lite produces $95\%$ \texttt{not\_found}, where its refutation is preserving almost all candidates rather than retiring them. Qwen3.5-VL-27B sits in between at $35\%/62\%/4\%$. A backbone that rarely rejects candidates has a refutation score systematically inflated toward anomaly, a runtime-observable signal that the agent's score scale on this backbone is miscalibrated and useful for deployment triage even without test labels.

\section{Leave-one-domain-out sensitivity (full table)}
\label{app:loo}

\begin{table}[h]
\centering
\caption{Leave-one-domain-out macro AUROC gain of score fusion over. All 11 LOO splits are positive on every backbone.}
\label{tab:loo}
\small
\begin{tabular}{lccc}
\toprule
Metric (pp) & GPT-5.5 & SeedVL & Qwen3.5 \\
\midrule
Full $\Delta$ (all 11 domains) & $+$1.4 & $+$1.9 & $+$5.9 \\
LOO min                        & $+$0.9 & $+$1.3 & $+$4.9 \\
LOO max                        & $+$2.2 & $+$3.1 & $+$7.0 \\
LOO std                        & 0.33   & 0.45   & 0.55   \\
Positive splits / 11           & 11 / 11 & 11 / 11 & 11 / 11 \\
\bottomrule
\end{tabular}
\end{table}

\section{Paired bootstrap (full nine-row table) and budget-matched controls}
\label{app:bootstrap}

Table~\ref{tab:main_results} in the main body lists the three GPT-5.5 rows; Table~\ref{tab:bootstrap_full} below includes all nine pairwise comparisons across the three backbones. Table~\ref{tab:budget} reports the negative-control budget-matched baselines that we ran on GPT-5.5 and SeedVL calibration splits.

\begin{table}[h]
\centering
\caption{Full stratified paired bootstrap table ($1{,}000$ resamples, per-domain stratification, $n{=}1298$).}
\label{tab:bootstrap_full}
\small
\begin{tabular}{llccc}
\toprule
Backbone & Comparison & $\Delta$ & 95\% CI & sig.\ at $p{<}0.05$ \\
\midrule
\multirow{3}{*}{GPT-5.5}  & fusion vs         & $+0.014$ & $[+0.003, +0.025]$ & \checkmark \\
                           & agent  vs         & $+0.005$ & $[-0.007, +0.015]$ & --- \\
                           & fusion vs agent     & $+0.009$ & $[-0.004, +0.022]$ & --- \\
\midrule
\multirow{3}{*}{SeedVL}   & fusion vs         & $+0.019$ & $[+0.008, +0.031]$ & \checkmark \\
                           & agent  vs         & $+0.023$ & $[+0.007, +0.039]$ & \checkmark \\
                           & fusion vs agent     & $-0.005$ & $[-0.019, +0.010]$ & --- \\
\midrule
\multirow{3}{*}{Qwen3.5}  & fusion vs         & $+0.059$ & $[+0.045, +0.076]$ & \checkmark \\
                           & agent  vs         & $+0.019$ & $[+0.009, +0.030]$ & \checkmark \\
                           & fusion vs agent     & $+0.040$ & $[+0.026, +0.055]$ & \checkmark \\
\bottomrule
\end{tabular}
\end{table}

\begin{table}[h]
\centering
\caption{Budget-matched negative controls (legacy-era numbers; included as historical record of disqualified designs and \emph{not comparable} with the ensemble numbers in Table~\ref{tab:main_results}). The Direct / SubspaceAD fusion / AnomalyClaw rows in this table use our older descriptor Direct and our older \texttt{AnomalyClaw} router system; the main-paper headline numbers (GPT-5.5 $0.864$, SeedVL $0.809$, Qwen3.5 $0.814$) are from the refutation agent$+$Direct ensemble on the same test split. ``Random escalation'' and ``DINOv2 blend'' are calibration-split numbers for disqualified designs. This table's purpose is only to show that blind second-call escalation and weaker expert blends are negative controls; it should not be read as a competing test-split comparison against our headline recipe.}
\label{tab:budget}
\small
\begin{tabular}{lcccc}
\toprule
Method & Calls/img & GPT-5.5 & SeedVL & Qwen3.5 \\
\midrule
Direct                                 & 1.00 & 0.822 & 0.795 & 0.792 \\
Random escalation 30\%                  & 1.30 & 0.795 & 0.750 & --- \\
Direct $+$ 0.2$\cdot$DINOv2 blend       & 1.00 & 0.830 & 0.807 & --- \\
\textbf{Direct $+$ 0.2$\cdot$SubspaceAD fusion} & \textbf{1.00} & \textbf{0.835} & 0.813 & \textbf{0.851} \\
\textbf{AnomalyClaw}             & $\sim$1.3 & 0.826 & \textbf{0.818} & 0.811 \\
\bottomrule
\end{tabular}
\end{table}

\section{VLM$\times$Expert complementarity matrix (full figure)}
\label{app:complementarity}

\begin{figure}[h]
\centering
\includegraphics[width=\linewidth]{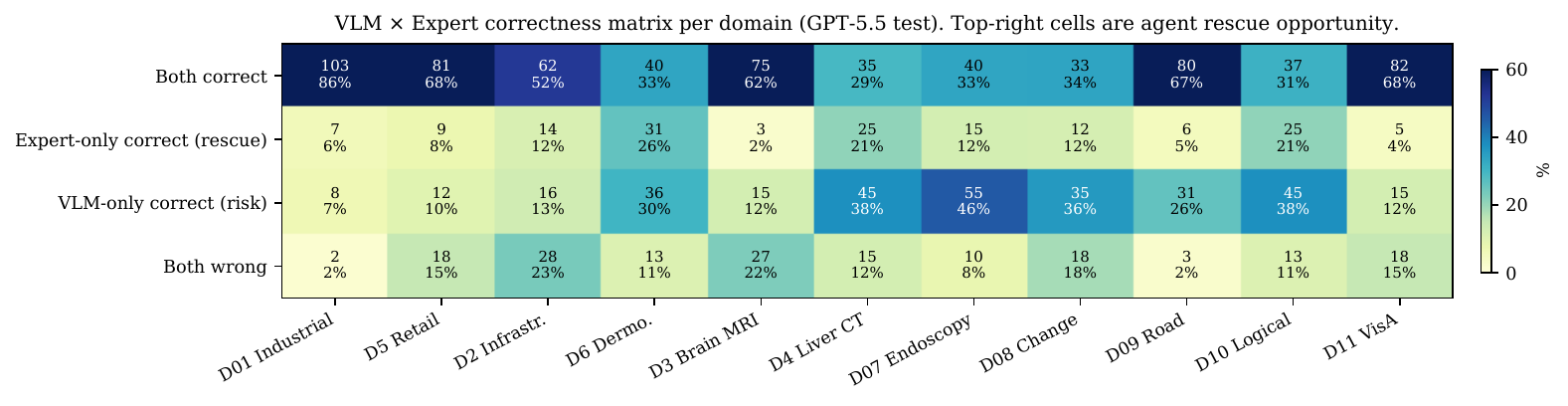}
\caption{VLM$\times$Expert correctness matrix per domain (GPT-5.5 test). Numbers are \% of items. ``Expert-only correct'' is the upper bound of what the agent can recover; ``VLM-only correct'' is what it risks losing by blindly trusting the expert.}
\label{fig:complementarity}
\end{figure}

\section{Legacy router architecture (pre-refutation)}
\label{app:legacy_router}

\emph{This section documents a prior router-based agent architecture (Routes A/B/C/D, descriptor router, hotspot cropper, proposer-advocate debate). The five subsections below are retained as reproducibility artifacts and do not describe the current refutation agent reported in \S\ref{sec:method} and Appendix~\ref{app:refutation}.}

\subsection{Decision trace example}
\label{app:trace}

For historical record, every prediction in the legacy router system carried a full trace that a human reviewer could audit. For illustration, we show two trace snippets from the actual test runs (Qwen3.5 backbone) --- first, the autonomous router plan generated for one domain, then a per-item agent execution.

\paragraph{Autonomous router plan (D10 BMAD-Liver, Qwen3.5).}
\begin{quote}\small
\textbf{Strategy:} interpret. \textbf{Expert:} subspacead. \textbf{Tools:} \{\texttt{domain\_descriptor}, \texttt{hotspot\_cropper}\}.\\[2pt]
\textbf{Reasoning (verbatim from Qwen3.5):} ``\emph{SubspaceAD is the designated expert for liver CT texture anomalies, and the interpret strategy provides a critical safety net to verify subtle focal lesions that the VLM may miss. The hotspot cropper makes the second-call evidence explicit when the expert flags a concentrated abnormality.}''\\[2pt]
\textbf{Cost:} 1 text-only VLM call (671 input / 1013 output tokens) --- amortized across all 120 D10 test items.
\end{quote}

\paragraph{Per-item agent execution (D1 Industrial, item \texttt{D1\_0063}, SeedVL).}
\begin{quote}\small
\textbf{Phase 1 (Plan lookup).} Cached plan for D1 = (\texttt{strategy:fusion}, \texttt{expert:subspacead}, tools \{\texttt{domain\_descriptor}, \texttt{hotspot\_cropper}\}).\\[2pt]
\textbf{Phase 2 (Expert).} \texttt{SubspaceAD.score} = 17.95 (global calibration median $m{=}40.0$); \texttt{top-5 patches} at $(22,24), (10,24), (10,22), (33,32), (\ldots)$ in the $48{\times}48$ grid. Derived $\rho = 159.7$, $\kappa = 1.43$.\\[2pt]
\textbf{Phase 3 (Online override).} $\hat{y}_{\mathrm{}}{=}$normal AND $\rho > 0.8$ $\Rightarrow$ promote strategy \emph{fusion} $\to$ \emph{interpret}.\\[2pt]
\textbf{Interpret call.} VLM re-examines the cropped hotspot bounding box (image coords $(307, 137)$--$(459, 478)$). \texttt{image\_label}=``anomalous'', \texttt{confidence}=$0.85$, \texttt{evidence}=``a linear scratch visible at the flagged region was missed at global inspection.''\\[2pt]
\textbf{Final decision.} anomaly; score $=$ 0.85. Ground truth: anomalous. Agent escalated correctly via the online expert override.
\end{quote}

Every test-split item has an analogous trace, serialized as JSON in the released \texttt{plan} and \texttt{raw\_output} fields. Score fusion, by contrast, produces only a scalar.

\subsection{Route C (Component Enumeration) prototype}
\label{app:routec}

We prototyped a fourth route, \emph{Component Enumeration} (Route C), for logical anomalies (D10 MVTec-LOCO; legacy display code, current taxonomy: D3), where the anomaly is defined by a wrong count or a wrong spatial arrangement of components rather than by a texture defect. Route C triggers when $\rho{>}1.2$ and $\kappa{<}1.10$ (strong but \emph{dispersed} expert signal). The action is a second VLM call with a specialized enumerate prompt that asks the model to list components in the query, rather than the references, and to report count differences.

On D10 (legacy code) calibration (20 items, 4 reference images per query), Route C reaches 60\% accuracy versus 55\%. The gain is modest for two reasons: (i) VLMs struggle to count small, similar objects reliably, and (ii) the 20-item calibration split yields noisy threshold estimates. Route C is \emph{not} part of the main-table AnomalyClaw variant reported in Table~\ref{tab:main_results}: the reported numbers use only Routes A/B/D. We include Route C here as a concrete direction for future work and flag it as a prototype rather than a validated component of the agent.

\subsection{Routing distribution}
\label{app:routing}

\begin{figure}[h]
\centering
\includegraphics[width=1.0\linewidth]{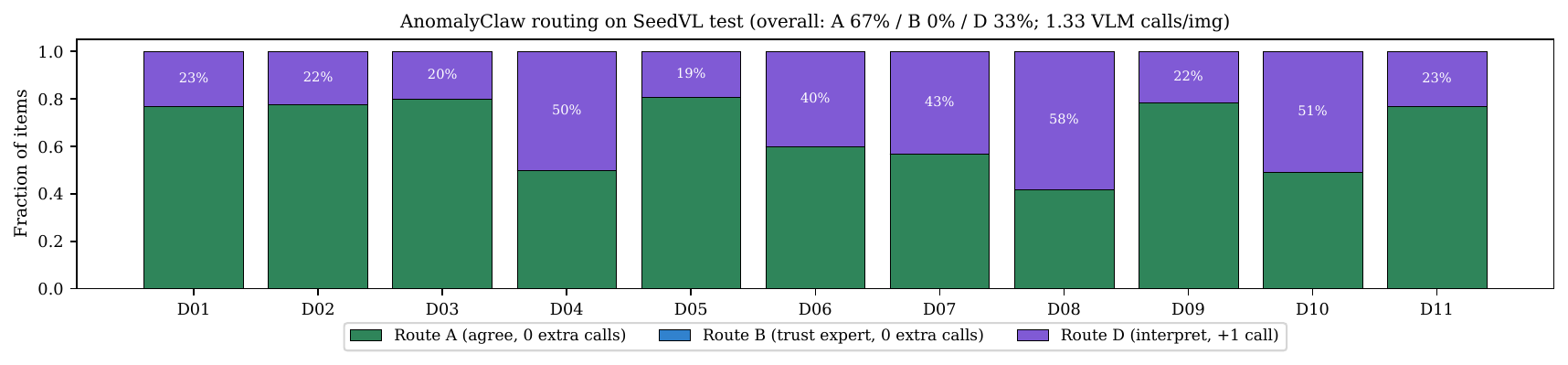}
\caption{AnomalyClaw routing distribution on the SeedVL test split. 67\% of items resolve by Route A (VLM and expert agree, 0 extra calls); Route D (interpret) is invoked on 33\% of items; Route B (trust expert) does not fire on any test item because the asymmetric anomaly-trust gate handles strong concentrated signals before they reach Route B.}
\label{fig:routing_appendix}
\end{figure}

\subsection{Proposer--advocate debate protocol}
\label{app:debate}

The \emph{debate} strategy runs two VLM calls in sequence. The Proposer receives the query, references, and the task-anchored descriptor, and returns a JSON object \texttt{\{image\_label, anomaly\_type, evidence, confidence, bbox\}}. The Advocate receives the Proposer's output verbatim, along with an adversarial prompt that asks it to refute the Proposer's claim using the same references, and returns a JSON object \texttt{\{refute\_confidence, counter\_evidence\}}. We combine both into a final score via the rule: if \texttt{refute\_confidence}\,$\geq 0.6$, overturn the Proposer; if \texttt{proposer.confidence}\,$\geq 0.6$ and \texttt{refute\_confidence}\,$\leq 0.4$, commit to the Proposer; otherwise take the Proposer's score unchanged. No third VLM call and no oracle. Temperature is 0 for both calls; the same descriptor $\phi_d$ is used.

\subsection{Descriptor router rules}
\label{app:router_rules}

The descriptor router maps each domain family to a default strategy. Rules were frozen once from the benchmark descriptor before any test query; no calibration AUROC was consulted.

\begin{table}[h]
\centering
\caption{Descriptor router rules.}
\label{tab:router_rules}
\small
\resizebox{\linewidth}{!}{%
\begin{tabular}{llp{7cm}}
\toprule
Family & Strategy & Reasoning \\
\midrule
industrial / retail / industrial\_visa & fusion & texture-dominant few-shot; expert is reliably discriminative, VLM calibrates \\
infrastructure & fusion & concrete cracks flagged by patch expert; descriptor narrows false positives \\
logical (D3 MVTec-LOCO) & fusion & component structure visible to patch expert; VLM rules out semantic look-alikes \\
dermoscopy (D8) & fusion & patch expert catches localised lesions; VLM rules out imaging artefacts \\
medical\_mixed / liver\_ct (D10) / brain\_mri (D9) & fusion & SubspaceAD complements VLM on non-trivial intensity shifts \\
gi\_endoscopy (D11) & direct & VLM-strong: mucosa color/shape semantics beyond generic feature distance \\
remote-sensing change (D7 LEVIR) & direct & change semantics require descriptor-level task anchoring \\
road (D12 BDD+RoadAnomaly) & direct & semantic obstacle recognition dominates; expert is a distractor \\
\bottomrule
\end{tabular}%
}
\end{table}

\section{Tool contracts}
\label{app:tools}

Each tool has a small Python contract. Below are the full signatures; the implementations (Python/NumPy) are in \texttt{refine-logs/anomaclaw\_v2/registry.py}.

\begin{itemize}
    \item \texttt{domain\_descriptor(domain\_code) $\to$ str}: returns the task-anchored descriptor. Zero cost, no side effects.
    \item \texttt{reference\_retriever(query, refs, k=3) $\to$ list[int]}: cosine similarity of DINOv2-CLS tokens; returns top-$k$ reference indices.
    \item \texttt{hotspot\_cropper(query, patches, k=5, pad=0.15) $\to$ bbox}: takes the top-$k$ expert patches, computes a padded bounding box in image coordinates, and returns a crop for the VLM.
    \item \texttt{component\_counter(hotspot\_map, threshold=0.5) $\to$ (n, stats)}: connected-component labelling with skimage; returns component count and per-component area/centroid.
    \item \texttt{knowledge\_lookup(domain\_code) $\to$ list[str]}: returns a fixed, per-domain keyword list (length $\le 6$; written once during benchmark setup).
\end{itemize}

Tools are composed explicitly by each strategy: \emph{direct} uses only the descriptor; \emph{fusion} uses descriptor and expert scores; \emph{interpret} additionally uses the cropper and expert hotspots; \emph{debate} uses the descriptor and optionally the retriever. No tool is invoked outside its strategy's declared contract.

\section{Rulebook Pilot: L1 invariants + L2 oracle corrective rules}
\label{app:rulebook_pilot}

The domain-rule corpus reused by the verbalized injection extension
(\S\ref{ssec:osr_results}, +\textsc{Cluster} mode) was produced by an
earlier pilot that we ran on the same
CrossDomainVAD-12 dev/test split. We document it here for reproducibility.
The pilot's standalone results are not part of the main claim but its
artifacts (per-domain JSON rulebooks) are the \texttt{invariant},
\texttt{corrective\_fn}, and \texttt{corrective\_fp} entries that the verbalized rule-injection extension
rulebook RAG retrieves.

\paragraph{L1 --- Reference-only invariants.}
For each (domain, category) pair, a reflector VLM (Qwen3.5-VL-27B) is shown
$n{=}8$ normal reference images and asked to list visual \emph{invariants}
--- concrete properties verifiable in \emph{every} reference --- restricted
to six types:
\begin{itemize}[topsep=1pt,parsep=0pt,itemsep=1pt]
  \item \texttt{count}: exact number of parts, segments, or discrete items.
  \item \texttt{symmetry}: axes or symmetries preserved across references.
  \item \texttt{spatial\_layout}: global arrangement of components.
  \item \texttt{color\_palette}: colour range observed across refs.
  \item \texttt{texture}: surface or fabric texture properties.
  \item \texttt{structural}: overall shape, topology, or part connectivity.
\end{itemize}
The reflector is forbidden from listing ``hypothesized anomaly modes'' (an
earlier pilot design that collapsed into generic VLM-prior recitation).
Empty lists are allowed when no invariant can be verified across all
references.

\paragraph{L2 --- Oracle cluster rules.}
For each domain, we run Passive the refutation agent on the $40$-item dev split, form a
balanced batch of $K/2$ false negatives (gt$=$1, $s{<}0.5$) plus $K/2$
false positives (gt$=$0, $s{\ge}0.5$) with $K{=}10$, reveal their manifest
labels, and ask a reflector VLM to propose $1$--$3$ \emph{corrective}
rules per side over the whole batch at once. Per-item rules are forbidden
(they over-generate); \texttt{normal\_tolerance} rules must name a specific
visual feature rather than blanket-suppress an anomaly class.

\paragraph{Rule store.}
For each domain, we emit one JSON file containing the invariants and
correctives tagged with metadata:
$(\mathrm{type},\ \mathrm{category}\ [\text{or null}],\ \mathrm{text},\ \mathrm{source\_items},\ \mathrm{confidence})$.
The agent never consumes these rules directly; instead, it reads
RAG-retrieved top-$k$ subsets filtered by the query's $(\mathrm{domain},
\mathrm{category})$ metadata (\S\ref{ssec:osr}). The stacked store
size ranges from 10 rules (D12 road safety, single category with few
invariants) to 86 rules (D1 MVTec-AD, eight categories each contributing
count/color/texture/structural invariants).

\paragraph{Artefacts.}
Per-domain rulebooks (Passive baseline, \textsc{+Cluster}, and \textsc{OSR}) and the rule-injection trajectory logs are released with the code.

\section{Beyond AUROC: deployment-relevant metrics}
\label{app:beyond_auroc}

AUROC is a threshold-free ranking metric, and the headline numbers in Table~\ref{tab:main_results} are reported as macro AUROC for direct comparability with prior cross-domain VAD work. AUROC alone, however, undersells what matters at deployment in medical, road-safety, or infrastructure settings: false-positive rate at high recall, threshold transfer, calibration, and image- vs.\ region-level localization are all separate axes a model can succeed or fail on independently of its AUROC. As an exploratory complement to Table~\ref{tab:main_results}, Table~\ref{tab:beyond_auroc} reports per-domain AUPRC and FPR@95TPR on the SeedVL backbone, on the same agent run.

\begin{table}[h]
\centering
\caption{Beyond-AUROC per-domain metrics on the \textbf{SeedVL} backbone, $n{=}1{,}418$ test items. AUPRC under prevalence baseline $\pi$; FPR@95TPR is the fraction of normal items mis-flagged when the threshold is set so that recall on anomalies reaches $95\%$ (lower is better). Reported on the same agent run as Table~\ref{tab:main_results}; per-domain AUROC matches that table to $\le 0.001$.}
\label{tab:beyond_auroc}
\setlength{\tabcolsep}{3.5pt}
\begin{tabular}{l|cc|cc|cc}
\toprule
 & \multicolumn{2}{c|}{AUROC} & \multicolumn{2}{c|}{AUPRC} & \multicolumn{2}{c}{FPR@95TPR ($\downarrow$)} \\
Domain & Direct & \textbf{Ours} & Direct & \textbf{Ours} & Direct & \textbf{Ours} \\
\midrule
D1 MVTec-AD       & 0.841 & \textbf{0.891} & 0.772 & \textbf{0.846} & 0.550 & \textbf{0.350} \\
D2 VisA           & 0.889 & \textbf{0.932} & 0.853 & \textbf{0.928} & 0.483 & \textbf{0.233} \\
D3 MVTec-LOCO     & 0.683 & \textbf{0.819} & 0.603 & \textbf{0.765} & 0.683 & \textbf{0.517} \\
D4 Real3D-AD      & 0.530 & \textbf{0.536} & 0.525 & \textbf{0.530} & 0.950 & \textbf{0.933} \\
D5 GoodsAD        & 0.872 & \textbf{0.878} & 0.844 & \textbf{0.864} & 0.900 & \textbf{0.533} \\
D6 SDNET          & 0.445 & \textbf{0.706} & 0.465 & \textbf{0.701} & 0.933 & \textbf{0.700} \\
D7 LEVIR          & 0.754 & \textbf{0.799} & 0.754 & \textbf{0.784} & 0.837 & \textbf{0.816} \\
D8 Derma          & 0.630 & \textbf{0.673} & 0.582 & \textbf{0.638} & 0.900 & \textbf{0.867} \\
D9 BraTS          & 0.616 & \textbf{0.685} & 0.561 & \textbf{0.619} & \textbf{0.717} & 0.750 \\
D10 Liver CT      & 0.465 & \textbf{0.560} & 0.480 & \textbf{0.522} & 1.000 & \textbf{0.767} \\
D11 Kvasir        & 0.731 & \textbf{0.759} & 0.697 & \textbf{0.748} & 0.867 & \textbf{0.683} \\
D12 Road          & 0.798 & \textbf{0.969} & 0.724 & \textbf{0.961} & 0.517 & \textbf{0.167} \\
\midrule
\textbf{Macro}    & 0.688 & \textbf{0.767} & 0.655 & \textbf{0.742} & 0.778 & \textbf{0.610} \\
$\Delta$ vs Direct &       & $+7.93$       &       & $+8.72$       &       & $-16.84$ \\
\bottomrule
\end{tabular}%
\end{table}

The picture is qualitatively the same as macro AUROC: AUPRC tracks AUROC closely (correlation across $12$ domains $\rho{=}0.93$), and FPR@95TPR drops on $11/12$ domains, with the largest reductions on D5 GoodsAD ($-37$\,pp), D12 Road ($-35$\,pp), and D2 VisA ($-25$\,pp). The single regression on FPR@95TPR is D9 BraTS ($+3$\,pp), where AUROC improves but the TPR=$0.95$ operating point trades a slight FP increase for the recall lift; D4 Real3D-AD shows essentially no separation on either branch, consistent with its near-chance AUROC.

\paragraph{Limitations of this section.}
The full deployment-grade evaluation we believe this benchmark eventually deserves --- per-backbone AUPRC, FPR@95TPR, ECE, Brier score, image-level vs.\ pixel-level localization, calibrated-threshold precision/recall, threshold-transfer between calibration and test, and a typology of failure modes (false-positive categories, refutation-induced over-clearance, etc.) --- is beyond the scope of this submission. We report SeedVL here as a representative case because its raw item-level scores match the Table~\ref{tab:main_results} aggregates exactly; the released artifact will accompany the full per-backbone deployment metric panel. Throughout the main paper, we use the language ``graded anomaly score'' rather than ``calibrated score'' to avoid overclaiming what an unaltered VLM-confidence readout, even after fusion, can guarantee.

\section{Tool provenance and leakage controls}
\label{app:tool_provenance}

A reasonable concern with any tool-using VLM agent is that gains may come from tool priors or from data leakage rather than from VLM reasoning. We list every tool's training data, what test-time information it sees, and whether it could leak.

\paragraph{Tools that have a learned component.}
\begin{itemize}\setlength\itemsep{1pt}
\item \texttt{expert\_score} (L3) wraps a frozen \textbf{SubspaceAD}~\citep{subspacead2026} pipeline. SubspaceAD itself is training-free in the AD sense: it takes a few-shot pool of normal references at query time, extracts patch tokens with a frozen DINOv2 backbone~\citep{oquab2024dinov2} pre-trained on LVD-142M (a curated $142$M-image internet corpus, with no exposure to MVTec-AD, VisA, MVTec-LOCO, BraTS, BMAD, or any of the other CrossDomainVAD-12 sources), fits a per-query PCA subspace on those reference tokens, and scores the query by the distance of its tokens from that subspace. No anomalies are observed at any stage, and no per-domain weights are fine-tuned.
\item \texttt{reference\_retriever} (L4) ranks normal references for the current query by cosine similarity in the same frozen DINOv2 feature space. It does not see anomaly labels.
\end{itemize}

\paragraph{Tools that are pure image processing.}
\texttt{side\_by\_side} (L2), \texttt{zoom\_bbox} (L5), \texttt{image\_diff}, \texttt{rotate\_align}, \texttt{segment\_and\_count}, \texttt{component\_counter}, \texttt{texture\_fft}, and \texttt{patch\_grid} are deterministic image-processing primitives that do not have a learned component; they take a query and references and return crops, diffs, edges, FFT spectra, etc.

\paragraph{What test-time information the agent does and does not see.}
At inference the VLM sees: the query image, $1$--$10$ normal reference images for the same category/scene, a single one-sentence task descriptor for the domain (Appendix~\ref{app:benchmark}), and the results of any tools it invokes during the $K{=}5$-turn trajectory. It does not see: per-image anomaly labels, the test-set list, dataset-wide normal/anomaly statistics, category-name lists revealing the test split, or any domain-specific prior fitted on test split items. The per-domain hint in the turn-1 user message names the domain by family (e.g., ``infrastructure'') and recommends a tool pipeline; it does not contain anomaly statistics or label distributions.

\paragraph{Calibration vs.\ test discipline.}
The $20{\times}12{=}240$-item calibration split is held out from the $1{,}418$-item test split with no overlap. Calibration is used (a) to pre-fit the cached \texttt{expert\_score} reliability flag per domain (a Boolean ``available / not available'' decision based on whether SubspaceAD's calibration AUROC exceeds chance) and (b) for the prompt-engineering choices reported in Appendix~\ref{app:descriptor}. No agent component is fit on test items, no test-derived statistic conditions any prompt, and the same $13$-tool catalog with the same applicability gates is used identically across all three VLM backbones.

\paragraph{Cross-backbone tool consistency.}
The agent code dispatches the same Python tool implementations regardless of which VLM is at the wheel. The differences across backbones reported in Appendix~\ref{app:behavior} (turn-1 share, verdict mix, tool diversity) are downstream of the VLM's free-text choices over a fixed catalog, not of any tool definition or per-backbone tool weighting.

\paragraph{What this section does \emph{not} rule out.}
DINOv2 is pre-trained on LVD-142M, which itself contains internet imagery. We cannot rule out that some images in CrossDomainVAD-12 source datasets (e.g., MVTec-AD, which is widely distributed open data) are embedded in DINOv2's pre-training distribution at the level of statistical similarity. This is a property shared with virtually every published VLM-based AD method that uses a public foundation model; we flag it as a limitation rather than a clean training-free guarantee.

\section{Refutation protocol details}
\label{app:refutation}

The refutation agent's full JSON schema (suspect-list, refutation-target, verdict-and-update), tool contracts, task preamble builder, and per-backbone prompt templates are released with the code. Figure~\ref{fig:architecture} (main body) gives the visual overview; the structured output schema distinguishes the three turn phases via the \texttt{action} field (\texttt{call\_tool} or \texttt{final}) and the three verdict categories (\texttt{found\_in\_ref} / \texttt{not\_found} / \texttt{inconclusive}).

\section{Cross-benchmark transfer: MMAD per-dataset}
\label{app:mmad}

On a stratified $500$-image sample of MMAD (seed $42$, $n{=}483$ AD questions across DS-MVTec, MVTec-LOCO, VisA, GoodsAD), the Qwen3.5-VL-27B ensemble reaches AUROC $0.79$ vs Direct $0.76$ with the same $0.5{:}0.5$ blend used in Table~\ref{tab:main_results}. No retuning. The gain is concentrated on GoodsAD and MVTec-LOCO, matching the per-domain pattern on CrossDomainVAD-12 (Table~\ref{tab:main_results}).

\section{Verbalized injection: rulebook pipeline}
\label{app:rulebook_pipeline}

\paragraph{\textsc{+Cluster} (offline, $K{=}10$ oracle labels/domain).} The cluster rule store is built offline in two stages on the dev split: (A) per-domain Passive-ensemble evaluation on the $20$-item selection subset $\mathcal{D}_{\mathrm{sel}}$, then (B) cluster reflection on a balanced $K/2{+}K/2$ FN/FP batch with $K{=}10$ revealed labels per domain. At inference, the (domain, category) rule subset is retrieved by metadata filter, top-$3$ rules are concatenated with the task anchor and prepended as user-message context to \emph{both} the Direct branch's prompt and the refutation branch's turn-$1$ user message; the \S4 ensemble blend is unchanged.

\paragraph{\textsc{OSR} (online, zero oracle).} The OSR rulebook starts empty at the beginning of the test pass and grows online as described in \S\ref{ssec:osr}. The reflector VLM is the same Qwen3.5-VL-27B / GPT-5.5 used as the agent backbone; its prompt receives the full disagreement batch (queries, references, both branches' trajectories, candidate features) and emits at most $3$ rules with self-assigned confidence in $\{\text{low}, \text{medium}, \text{high}\}$. Rules at $\ge$\, medium confidence are appended to the per-domain rule list with metadata $(\mathrm{type}, \mathrm{category}, \mathrm{text}, \mathrm{source\_items}, \mathrm{confidence})$; subsequent items are retrieved via the same metadata filter as Cluster.

\paragraph{$\alpha$-tune baseline.} For each backbone we draw a stratified $K{=}10$-label sample per domain from $\mathcal{D}_{\mathrm{val}}$ and select $\alpha_d \in \{0.05, 0.10, \ldots, 0.95\}$ that maximises domain AUROC; the resulting $\alpha_d$ is applied at test time. Headline values use a fixed sample seed ($\textit{seed}{=}0$) for determinism, matching the protocol of \textsc{+Cluster}. Variance over $30$ resamples is high at $K{=}10$/domain ($\alpha_d^{*}$ range $[0.05, 0.95]$, test macro range $\sim 0.005$); we report the seed-$0$ deterministic instance to keep the budget protocol identical to \textsc{+Cluster}, which is the apples-to-apples comparison.

\section{OSR per-domain breakdown}
\label{app:osr_per_domain}

Table~\ref{tab:osr_per_domain} reports per-domain AUROC for the four verbalized modes on Qwen3.5-VL-27B and per-domain rule counts accumulated by \textsc{OSR} during the test pass. \textsc{OSR} wins or ties on $5$ of $12$ domains and accumulates the most rules where the \S\ref{sec:main} ensemble shows the most internal disagreement (D10 $9$, D3/D8 $8$, D11 $7$). On D2 VisA and D9 BraTS, the disagreement rate is below the threshold for the $N{=}10$ queue to fill within the $120$-item test pass; those domains stay at \textsc{Passive} and no rules are written. On D8 DermaMNIST \textsc{OSR}'s rules \emph{hurt} ($-5.0$\,pp): the disagreement signal flipped to favor Direct (which is ceiling-strong on DermaMNIST), but the reflector encoded the refutation pattern, mis-routing items at test time --- the known failure mode of zero-oracle reflection on backbone-saturated domains.

\begin{table}[t]
\centering
\caption{\textbf{Per-domain AUROC on CrossDomainVAD-12 test, Qwen3.5-VL-27B} ($\alpha{=}0.5$ throughout). \textsc{rules} = number of rules \textsc{OSR} accumulated for that domain by end of test pass; best per row in \textbf{bold}.}
\label{tab:osr_per_domain}
\small
\setlength{\tabcolsep}{3pt}
\begin{tabular}{llccccc}
\toprule
Dom & Source / defect type & \textsc{Passive} & +\textsc{Anchor} & +\textsc{Cluster} & \textbf{\textsc{OSR}} & rules \\
\midrule
D1  & MVTec-AD / surface defect   & 0.956 & 0.955 & 0.928 & \textbf{0.964} & 2 \\
D2  & VisA / fine-grained defect  & 0.772 & \textbf{0.845} & \textbf{0.845} & 0.839 & 0 \\
D3  & MVTec-LOCO / logical        & 0.673 & 0.680 & 0.699 & \textbf{0.701} & 8 \\
D4  & Real3D-AD / geometric       & 0.517 & \textbf{0.535} & 0.515 & 0.525 & 3 \\
D5  & GoodsAD / packaging         & \textbf{0.649} & 0.614 & 0.623 & 0.624 & 3 \\
D6  & SDNET / concrete crack      & 0.672 & 0.839 & 0.764 & \textbf{0.869} & 6 \\
D7  & LEVIR / building change     & 0.842 & 0.844 & \textbf{0.896} & 0.859 & 5 \\
D8  & DermaMNIST / melanoma       & 0.630 & \textbf{0.698} & 0.646 & 0.580 & 8 \\
D9  & BraTS / tumor               & 0.841 & \textbf{0.867} & 0.863 & 0.860 & 0 \\
D10 & BMAD-Liver / pathology      & 0.616 & 0.624 & \textbf{0.673} & 0.654 & 9 \\
D11 & HyperKvasir / polyp/ulcer   & \textbf{0.817} & 0.810 & 0.799 & 0.782 & 7 \\
D12 & BDD+RoadAnomaly / obstacle  & \textbf{0.990} & 0.966 & 0.964 & 0.970 & 5 \\
\midrule
\textbf{Macro} & & 0.7480 & 0.7731 & 0.7679 & $\mathbf{0.7689}$ & --- \\
\textbf{$\Delta$ vs Passive} & & --- & $+2.51$ & $+1.99$ & $\mathbf{+2.09}$ & --- \\
\textbf{$P(\Delta_{\text{boot}}{>}0)$} & & --- & $0.92$ & $0.89$ & $\mathbf{0.87}$ & --- \\
\bottomrule
\end{tabular}
\end{table}

On GPT-5.5 \textsc{OSR} accumulates $71$ rules across the $12$ domains (D8 $11$, D3/D9 $9$, D10 $8$, D4/D5 $6$, D2/D7 $5$, D1/D11/D12 $3$) and wins on D10 BMAD-Liver ($+10.9$\,pp), D6 SDNET ($+4.4$\,pp), D3 MVTec-LOCO ($+3.3$\,pp); loses on D8 DermaMNIST ($-6.4$\,pp, the same domain that hurt on Qwen3.5) and D11 Kvasir ($-3.1$\,pp). The mechanism is alive on the strong backbone too; the limit is residual headroom (\textsc{Passive} is already $0.8135$).

\newpage

\end{document}